%% file: paper.tex
\documentclass[acmtog,nonacm]{acmart}
\usepackage{booktabs} 
\usepackage{multirow}
\usepackage{tabularx}
\usepackage{wrapfig}
\usepackage{mathtools}
\usepackage{lipsum}
\usepackage{footmisc}
\usepackage{comment}
\usepackage[ruled]{algorithm2e}

\SetAlFnt{\small}
\SetAlCapFnt{\small}
\SetAlCapNameFnt{\small}
\SetAlCapHSkip{0pt}

\setcopyright{none}
\acmDOI{}

\begin{document}
\def\negativevspace{}	

\newcommand{\para}[1]{\vspace{.05in}\noindent\textbf{#1}}
\def\ie{\emph{i.e.}}
\def\eg{\emph{e.g.}}
\def\etal{{\em et al.}}
\def\etc{{\em etc.}}
\newcolumntype{C}[1]{>{\centering\arraybackslash}p{#1}}
	
\title{Neural Wavelet-domain Diffusion for 3D Shape Generation, Inversion, and Manipulation}





\author{Jingyu Hu$^{*}$, Ka-Hei Hui$^{*}$, Zhengzhe Liu}
\affiliation{%
	\institution{The Chinese University of Hong Kong} \country{HK SAR, China}}
\author{Ruihui Li}
\affiliation{%
	\institution{Hunan University}\country{HK SAR, China}}
\author{Chi-Wing Fu}
\affiliation{%
	\institution{The Chinese University of Hong Kong}\country{HK SAR, China}}
\renewcommand\shortauthors{Hu et al.}

\input{teaser}
\input{abstract}

%
%


%
%


\maketitle

\input{introduction}
\input{rw}
\input{overview}
\input{method}
\input{evaluation}
\input{conclusion}

\bibliographystyle{ACM-Reference-Format}
\bibliography{bibliography}

\end{document}

%% file: teaser.tex
\begin{teaserfigure}
\centerline{\includegraphics[width=0.99\textwidth]{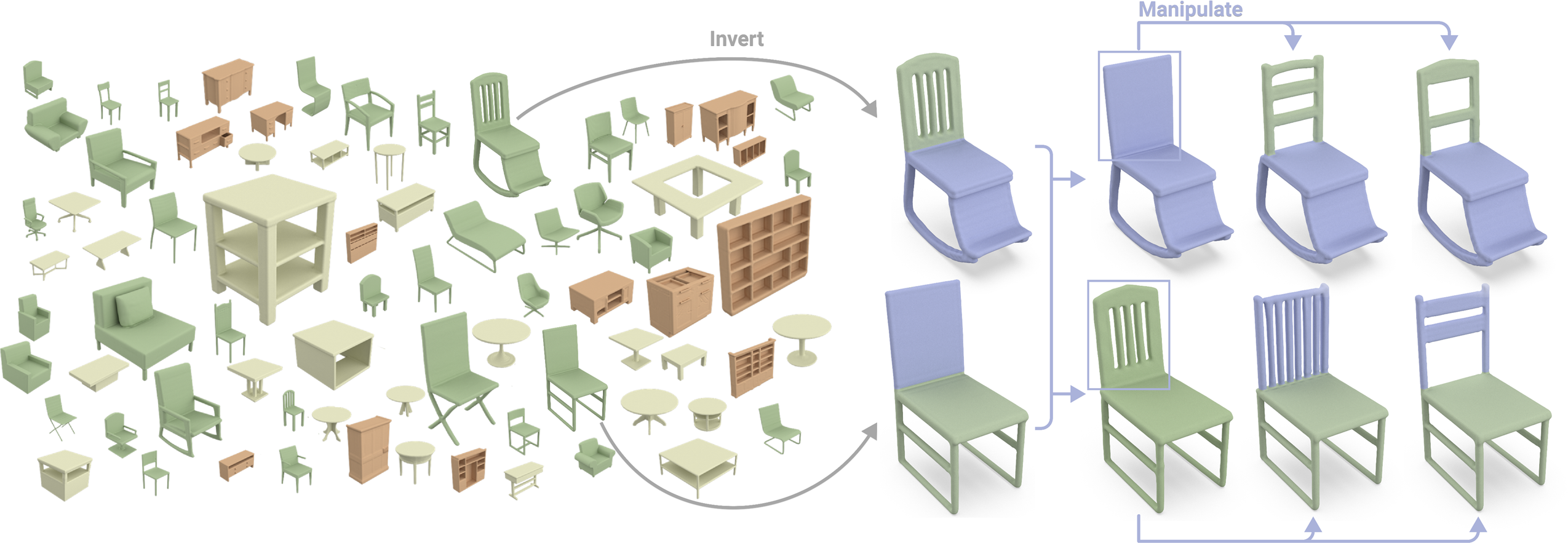}}
\vspace*{-3.5mm}
\caption{Our new framework is able to generate diverse and realistic 3D shapes that exhibit complex structures and topology, fine details, and clean surfaces, without obvious artifacts (left). 
Also, we can encode shapes and invert (reconstruct) them from their codes with high fidelity (middle).
Further, we may manipulate shapes in a region-aware manner and generate new shapes by composing/re-generating
parts from existing or randomly-generated shapes (right).}
\label{fig:teaser}
\end{teaserfigure}

%% file: abstract.tex
\begin{abstract}
This paper presents a new approach for 3D shape generation, inversion, and manipulation, through a direct generative modeling on a continuous implicit representation in wavelet domain.
Specifically, we propose a {\em compact wavelet representation\/} with a pair of coarse and detail coefficient volumes to implicitly represent 3D shapes via truncated signed distance functions and multi-scale biorthogonal wavelets.
Then, we design a pair of neural networks: 
a diffusion-based {\em generator} to produce diverse shapes in the form of the coarse coefficient volumes and
a {\em detail predictor\/} to produce compatible detail coefficient volumes for 
introducing fine structures and details.
Further, we may jointly train an 
{\em encoder network\/} to learn a latent space for inverting shapes, allowing us to enable a rich variety of whole-shape and region-aware shape manipulations.
Both quantitative and qualitative experimental results manifest the compelling shape generation, inversion, and manipulation capabilities of our approach over the state-of-the-art methods.
\end{abstract}

%% file: introduction.tex
\section{Introduction}
\label{sec:intro}
Generative modeling of 3D shapes enables rapid creation of 3D contents, enriching extensive applications across graphics, vision, and VR/AR.
With the emerging large-scale 3D datasets~\cite{chang2015shapenet}, data-driven shape generation has gained increasing attention.
However, it is still very challenging to learn to generate 3D shapes that are diverse, realistic, and novel, while promoting controllability on part- or region-aware shape manipulations with high fidelity.

Existing shape generation models are developed mainly for voxels~\cite{girdhar2016learning,zhu2017rethinking,yang2018learning}, point clouds~\cite{fan2017point,jiang2018gal,achlioptas2018learning}, and meshes~\cite{wang2018pixel2mesh,groueix2018papier,smith2019geometrics,tang2019skeleton}. 
Typically, these representations cannot handle
high resolutions or irregular topology, thus unlikely producing high-fidelity results.
In contrast, implicit functions~\cite{mescheder2019occupancy,park2019deepsdf,chen2019learning} show improved performance in surface reconstructions.
By representing a 3D shape as a level set of discrete volume or a continuous field, we can flexibly extract a mesh object of arbitrary topology at desired resolution.

Existing generative models such as GANs and normalizing flows have shown great success in generating point clouds and voxels.
Yet, they cannot effectively generate implicit functions.
To represent a surface in 3D, a large number of point samples are required, even though many nearby samples are redundant.
Taking the occupancy field for instance, only regions near the surface have varying data values, yet we need huge efforts to encode samples in constant and smoothly-varying regions.
Such representation non-compactness and redundancy 
demands a huge computational cost and 
hinders the learning efficiency on implicit surfaces.

To address these challenges, some methods attempt to sample in a pre-trained latent space built on the reconstruction task~\cite{chen2019learning,mescheder2019occupancy} or convert the generated implicits into point clouds or voxels for adversarial learning~\cite{kleineberg2020adversarial,luo2021surfgen}. 
However, these regularizations can only be indirectly applied to the generated implicit functions, so they are not able to ensure the generation of realistic objects.
Hence, the visual quality of the generated shapes often shows a significant gap, as compared with the 3D reconstruction results, and the diversity of their generated shapes is also quite limited. 

Further, to promote controllability in whole-shape or part-wise manipulation on implicit shapes is challenging. Some works encode implicit shapes into latent codes and perform shape manipulation by modifying the latent codes~\cite{park2019deepsdf, chen2019learning, mescheder2019occupancy}. 
To achieve part-wise controllability, 
some methods~\cite{hertz2022spaghetti, hao2020dualsdf} learn to decompose implicit shapes into a set of templates and manipulate corresponding templates of specific parts.
Yet, these methods struggle to invert (implicitly encode then reconstruct) shapes faithfully due to the redundancy in the sample-based implicit representation, which further hinders the shape manipulation quality.
Recently,~\cite{lin2022neuform} propose a method to manipulate implicit shapes by introducing extra part annotations, which enhance the quality of shape inversion and manipulation.
Yet, preparing part annotations on 3D shapes is tedious and costly.
Also, instead of refining the latent code at testing, their method overfits each unseen shape and the process is time-consuming,~\ie, around 25 minutes per shape.


This work introduces a new approach for 3D shape generation, inversion, and manipulation, enabling a direct generative modeling on a continuous implicit representation in the compact wavelet frequency domain.
Overall, we have six key contributions:
(i) a compact wavelet representation (a pair of coarse and detail coefficient volumes) based on the biorthogonal wavelets and truncated signed distance field to implicitly encode 3D shapes, facilitating effective learning for shape generation and inversion;
(ii) a generator network formulated based on the diffusion probabilistic model~\cite{sohl2015deep} to produce coarse coefficient volumes from random noise samples, promoting the generation of diverse 
and novel
shapes; 
(iii) a detail predictor network, formulated to produce compatible detail coefficients to enhance the generation of fine details; 
(iv) an encoder network, jointly trained with the generator 
to build a latent space for supporting shape inversion and manipulation;
(v) a shape-guided refinement scheme to enhance the shape inversion quality; and 
(vi) region-aware manipulation for implicitly editing object regions 
without additional part annotations.

In this work, with the generator and detail predictor,
we can flexibly generate diverse
and realistic
shapes that are not necessarily the same as the training shapes.
Further with the encoder network, we can embed a shape, not necessarily in the training set, into our compact latent space for shape inversion and manipulation.
As Figure~\ref{fig:teaser} shows, our generated shapes exhibit diverse topology, clean surfaces, sharp boundaries, and fine details, without obvious artifacts.
Fine details such as curved/thin beams, small pulleys, and complex cabinets 
are very challenging for the existing 3D generation approaches to synthesize. Besides, our method can faithfully invert and reconstruct randomly-generated shapes, as shown in the middle of Figure~\ref{fig:teaser}. Further, we support a rich variety of shape manipulations,~\eg, composing/re-generating parts from existing or randomly-generated shapes; see the right of Figure ~\ref{fig:teaser}.

This work extends~\cite{hui2022neural}, which was presented very recently in a conference.
The previous work~\cite{hui2022neural} focuses mainly on unconditional 3D shape generation.
In this extended version, we first expand~\cite{hui2022neural} with an encoder network and the shape-guided refinement scheme to build a latent space, enabling us to faithfully invert 3D shapes.
Besides, we design the region-aware manipulation procedure to support applications beyond shape interpolation,~\eg, part replacement, part-wise interpolation, and part-wise re-generation. 
Further, by utilizing the compact latent space, we explore the potential of our method for reconstructing implicit shapes from point clouds or images.
Also, we perform various new experiments to evaluate our framework on shape inversion and shape manipulation, and include more comparisons, including GET3D~\cite{gao2022get3d}, a very recent work on 3D shape generation.
Both quantitative and qualitative experimental results manifest the superiority of our method on 3D generation, inversion, and manipulation over the
state-of-the-art methods.


%% file: rw.tex
\vspace*{-1mm}
\section{Related Work}

\label{sec:rw}
\paragraph{3D reconstruction via implicit function.}
Recently, many methods leverage the flexibility 
of implicit surface for 3D reconstructions from 
voxels~\cite{mescheder2019occupancy,chen2019learning}, complete/partial point clouds~\cite{park2019deepsdf,Liu2021MLS,yan2022shapeformer}, and RGB images~\cite{xu2019disn,xu2020ladybird,li2021d2im,tang2021skeletonnet}.
On the other hand, besides ground-truth field values, various supervisions have been explored to train the generation of implicit surfaces,~\eg, multi-view images~\cite{liu2019learning,niemeyer2020differentiable} and unoriented point clouds~\cite{atzmon2020sal,gropp2020implicit,zhao2021sign}.
Yet, the task of 3D reconstruction focuses mainly on synthesizing a high-quality 3D shape that best matches the input.
So, it is fundamentally very different from the task of 3D shape generation, which aims to learn the shape distribution of a given set of shapes for generating diverse, high-quality, and possibly novel shapes accordingly.

\vspace*{-7pt}
\paragraph{3D shape generation via implicit function.}
Unlike the 3D reconstruction task, 3D shape generation has no fixed ground truth to supervise the generation of each shape sample.
Exploring efficient guidance for implicit surface generation is still an open problem.
Some works attempt to use the reconstruction task to first learn a latent embedding~\cite{mescheder2019occupancy,chen2019learning,hao2020dualsdf,ibing20213d} then generate new shapes by decoding 
codes sampled from the learned latent space.
Recently,~\cite{hertz2022spaghetti} learn a latent space with a Gaussian-mixture-based autodecoder for shape generation and manipulation.
Though these approaches ensure a simple training process, the generated shapes have limited diversity restricted by the pre-trained shape space.
Some other works attempt to convert implicit surfaces to some other representations,~\eg, voxels~\cite{kleineberg2020adversarial,zheng2022sdfstylegan}, point cloud~\cite{kleineberg2020adversarial}, and mesh~\cite{luo2021surfgen}, 
for applying adversarial training.
Yet, the conversion inevitably leads to information loss in the generated implicit surfaces, thus reducing the training efficiency and generation quality.

In this work, we propose a compact wavelet representation for modeling the implicit surface and learn to synthesize it with a diffusion model.
By this means, we can effectively learn to generate the implicit representation without a pre-trained latent space and a representation conversion.
The results in Section~\ref{subsec:shape_gen} also show that our new approach is capable of producing diversified shapes of high visual quality, exceeding the state-of-the-art methods.


\vspace*{-7pt}
\paragraph{3D shape generation via other representations.}
\cite{smith2017improved,wu2016learning} explore voxels, a natural grid-based extension of 2D image.
Yet, the methods learn mainly coarse structures and fail to produce fine details
due to memory restriction.
Some other methods explore point clouds via GAN~\cite{gal2020mrgan,hui2020progressive,li2021spgan}, flow-based models~\cite{kim2020softflow,cai2020learning}, and diffusion models~\cite{zhou20213d}.
Due to the discrete nature of point clouds, 3D meshes reconstructed from them often contain artifacts.
This work focuses on implicit surface generation, aiming at generating high-quality and diverse meshes with fine details and
overcoming the limitations of
the existing representations.

 \begin{figure*}[!t]
 	\centering
 	\includegraphics[width=1.0\linewidth]{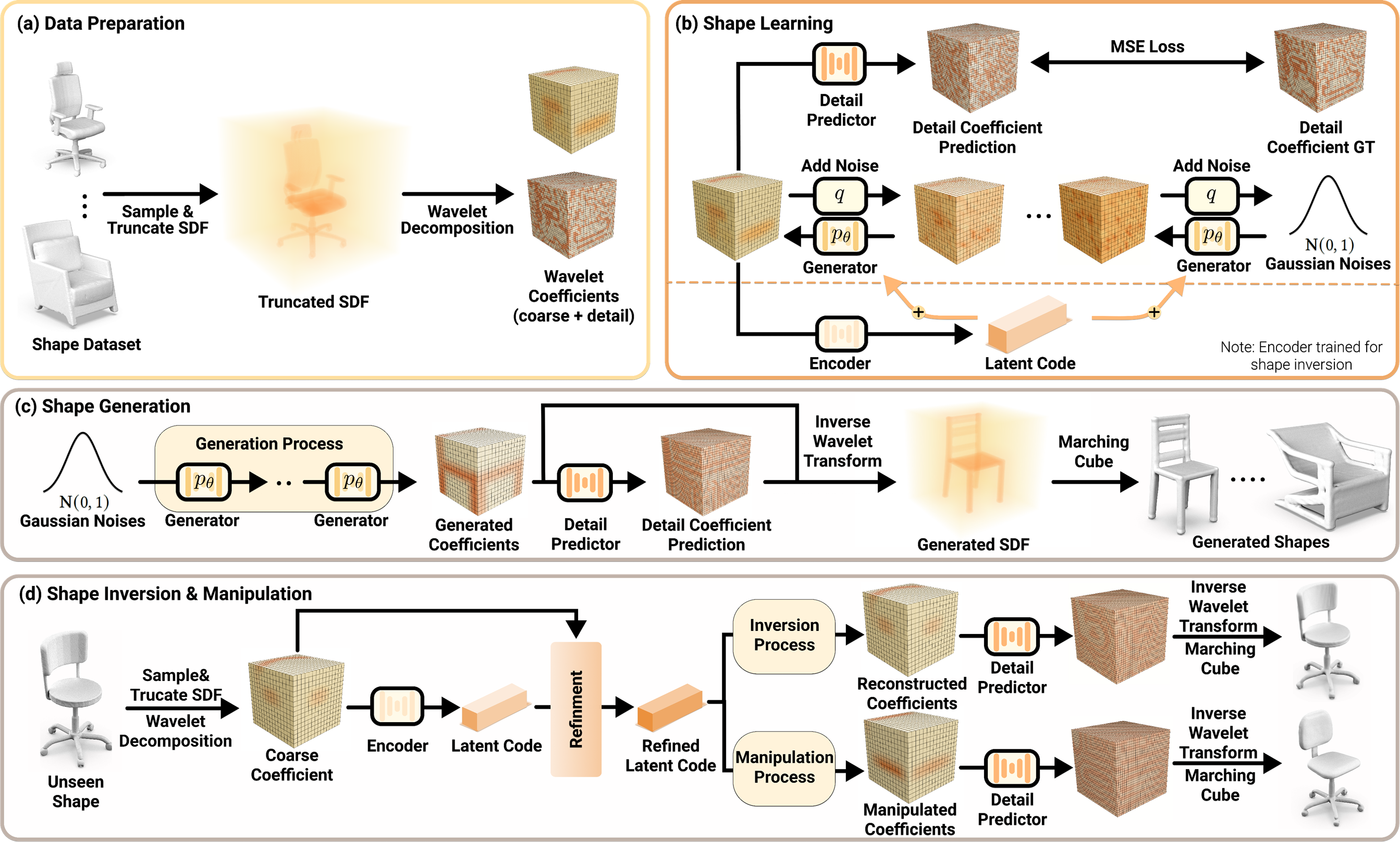}
 	\vspace*{-5mm}
 	\caption{Overview of our approach.
 	(a) {\em Data preparation\/} builds a compact wavelet representation (a pair of coarse and detail coefficient volumes)
 	for each input shape using a truncated signed distance field and a multi-scale wavelet decomposition.
 	(b) {\em Shape learning\/} trains the generator network to produce coarse coefficient volumes from random noise samples and trains the detail predictor network to produce detail coefficient volumes from coarse coefficient volumes. 
  Further, 
  an encoder is jointly trained with the generator to build a compact latent space for shape inversion and manipulation.
 	(c) {\em Shape generation\/} employs the trained generator to produce a coarse coefficient volume and then the trained detail predictor to further predict a compatible detail coefficient volume, followed by an inverse wavelet transform and marching cube, to generate the output 3D shape. 
(d) {\em Shape inversion \& manipulation\/}  employ the trained encoder to map a given shape to our compact latent space, then refine the latent code to enable faithful reconstruction and high-quality shape manipulation. 
}
	\label{fig:overview}
	\vspace*{-1.5mm}
\end{figure*}

\vspace*{-7pt}
\paragraph{3D shape inversion \& manipulation via implicit function}

The inversion task was first proposed for 2D images, aiming at embedding a given image, not necessarily in the training set, into a trained model's latent space~\cite{xia2022gan,bausemanticphoto,zhu2016generative,abdal2021styleflow,tewari2020pie,tov2021designing}, so that we can perform 
semantic
manipulations on the image and reconstruct a modified version of it from the manipulated latent code.
3D shape inversion is a relatively new topic.
So far, research works mainly explored the following 3D representations,~\eg, point clouds~\cite{zhang2021unsupervised}, voxels~\cite{wu2016learning}, 
and part-annotated bounding boxes or point clouds~\cite{mo2019structurenet, li2017grass}.

The existing inversion approaches for implicit representations can be divided into two categories: (i) leveraging an auto-encoder framework to map the given shape to the latent space~\cite{mescheder2019occupancy,chen2019learning, genova2020local} and (ii) optimizing the latent code at test time without using an additional encoder~\cite{park2019deepsdf, hao2020dualsdf, hertz2022spaghetti, lin2022neuform}.
We combine their strengths by initializing the latent code through an encoder network and further refining the code in a shape-guided manner; see Section~\ref{sec:result_inversion}
for the experimental results, which demonstrate the compelling performance of our method.

More importantly, 3D shape inversion enables user
manipulations on existing shapes in the compact latent space.
Some recent works~\cite{mo2019structurenet, gal2020mrgan, wei2020learning, li2021spgan, hui2022template} explore shape manipulation in the latent space via explicit 3D representations.
Yet, manipulation is still challenging for neural implicit representations.
First, some existing works~\cite{mescheder2019occupancy, chen2019learning, park2019deepsdf} 
lack part-level awareness in the manipulation.
Second, the inversion quality can be severely limited by the shape representative capability of their models~\cite{hertz2022spaghetti, hao2020dualsdf}.
A very recent work~\cite{lin2022neuform} enables 
part-aware
manipulation with good quality but it requires extra part annotations, which are costly to prepare.
In this work, we propose a rich variety of region-aware manipulations, besides whole-shape interpolation, without requiring part-level annotations; see Section~\ref{sec:result_manipulation} for 
the high-fidelity manipulated shapes that can be generated using our new approach.

\vspace*{-7pt}
\paragraph{Multi-scale neural implicit representation.}
This work also relates to multi-scale representations, so we discuss some 3D deep learning works in this area.
\cite{liu2020neural,takikawa2021neural,martel2021acorn,chibane2020implicit,chen2021multiresolution} predict multi-scale latent codes in an adaptive octree
to improve the reconstruction quality and inference efficiency.
\cite{fathony2020multiplicative} propose a band-limited network
to obtain a multi-scale representation by restricting the frequency magnitude of the basis functions.
Recently,~\cite{saragadam2022miner} adopt the Laplacian pyramid to extract multi-scale coefficients for multiple neural networks.
Unlike our work, this work overfits
each input object with an individual representation for efficient storage and rendering.
In contrast to our work on shape generation,
the above methods focus on improving 3D reconstruction performance by separately handling features at different levels.
In our work, we adopt a multi-scale implicit representation based on wavelets (motivated by~\cite{velho1994multiscale}) to build a compact representation
for high-quality shape generation.


\vspace*{-7pt}
\paragraph{Diffusion-based models.}
\cite{sohl2015deep,ho2020denoising,nichol2021improved,song2020denoising} 
recently show top performance in image generation, surpassing GAN-based models~\cite{dhariwal2021diffusion}. 
Recently,~\cite{luo2021diffusion,zhou20213d} adopt diffusion models for point cloud generation.
Yet, they fail to generate smooth surfaces and complex structures, as point clouds contain only discrete samples.
Distinctively, we adopt a compact representation based on wavelets to model the continuous signed distance field in our diffusion model, promoting 3D shape representation with
diverse structures and fine details.

%% file: overview.tex
\section{Overview}
\label{sec:overview}
As shown in Figure~\ref{fig:overview}, our approach has the following procedures:

\vspace*{4pt}
(i) {\em Data preparation}~is a one-time process for preparing a compact wavelet representation from each input shape; see Figure~\ref{fig:overview}(a).
For each shape, we sample a signed distance field (SDF) and 
truncate its distance values
to avoid redundant information.
Then, we transform the truncated SDF to the wavelet domain to produce a series of multi-scale coefficient volumes.
Importantly, we take {\em a pair of coarse and detail coefficient volumes\/} at the same scale as our compact wavelet representation for implicitly encoding the input shape.

\vspace*{4pt}
(ii) {\em Shape learning\/}~aims to train a pair of neural networks to learn the 3D shape distribution from the coarse and detail coefficient volumes; see Figure~\ref{fig:overview}(b).
First, we adopt the denoising diffusion probabilistic model~\cite{sohl2015deep} to formulate and train the {\em generator network\/} to learn to iteratively refine a random noise sample for generating diverse 3D shapes in the form of the coarse coefficient volume. 
Second, we design and train the {\em detail predictor network\/} to learn to produce the detail coefficient volume from the coarse coefficient volume for introducing further details in our generated shapes.
Besides, we jointly train an additional encoder with the generator for mapping the coarse coefficient volume to a compact latent code.
By doing so, the latent code can serve as a controllable condition in the generation, enabling applications such as shape inversion and manipulation; see (iv) below.

\vspace*{4pt}
(iii) {\em Shape generation\/}~employs the two trained networks to generate 3D shapes; see Figure~\ref{fig:overview}(c).
Starting from a random Gaussian noise sample, we first use the trained generator to produce the coarse coefficient volume then the detail predictor to produce an associated detail coefficient volume.
After that, we can perform an inverse wavelet transform, followed by the marching cube operator~\cite{lorensen1987marching}, to generate the output 
3D shape. 

\vspace*{4pt}
(iv) {\em Shape inversion \& manipulation\/}~aim to embed a given shape
in the generator's latent space, 
such that we may manipulate the shape and reconstruct it; see Figure~\ref{fig:overview}(d).
To invert a shape, we follow procedures (i) \& (ii) to sample a truncated signed distance field (TSDF) from the shape, transform it to the wavelet domain, and derive its latent code using the encoder jointly trained in procedure (ii).
Then, we enhance the correspondence between the shape and the latent code by refining the latent code via a back propagation with the frozen generator and encoder.
Further, we perform manipulation and reconstruction by feeding the derived (or manipulated) latent code as a condition on the generator to reconstruct the final shape.
Also, by utilizing the trained latent space,  
we can perform various region-aware manipulations on the given shape.

%% file: method.tex
\section{Method}
\label{sec:architecture}

\subsection{Compact Wavelet Representation}
\label{sec:data_preparation}
Preparing a compact wavelet representation of a given 3D shape (see Figure~\ref{fig:overview}(a)) involves the following two steps:
(i) implicitly represent the shape using a signed distance field (SDF); and
(ii) decompose the implicit representation via wavelet transform into coefficient volumes, each encoding a specific scale of the shape.

In the first step, we scale each shape to fit 
$[-0.9,+0.9]^3$ and sample an SDF of resolution $256^3$ to implicitly represent the shape.
Importantly, we truncate the distance p in the SDF to 
$[-0.1,+0.1]$, so regions not close to the
object surface are clipped to a constant.
We denote the truncated signed distance field (TSDF) for the $i$-th shape in training set as $S_i$.
Using $S_i$, we can significantly reduce the shape representation redundancy, such that the shape learning process can focus better on the shape structures and fine details.

The second step is a multi-scale wavelet decomposition~\cite{mallat1989theory,daubechies1990wavelet,velho1994multiscale} on the TSDF.
Here, we decompose $S_i$ into 
a high-frequency detail coefficient volume and 
a low-frequency coarse coefficient volume, which is roughly a compressed version of $S_i$.
We repeat this process $J$ times on the coarse coefficient volume of each scale, 
decomposing $S_i$ into a series of multi-level coefficient volumes.
We denote the coarse and detail coefficient volumes at the $j$-th step (scale) as $C^j_i$ and $D^j_i$, respectively, where $j = \{1,...,J\}$.
The representation is lossless, meaning that the extracted coefficient volumes together can faithfully
reconstruct the original TSDF via a series of inverse wavelet transforms.

\begin{figure}[!t]
	\centering
	\includegraphics[width=0.9\linewidth]{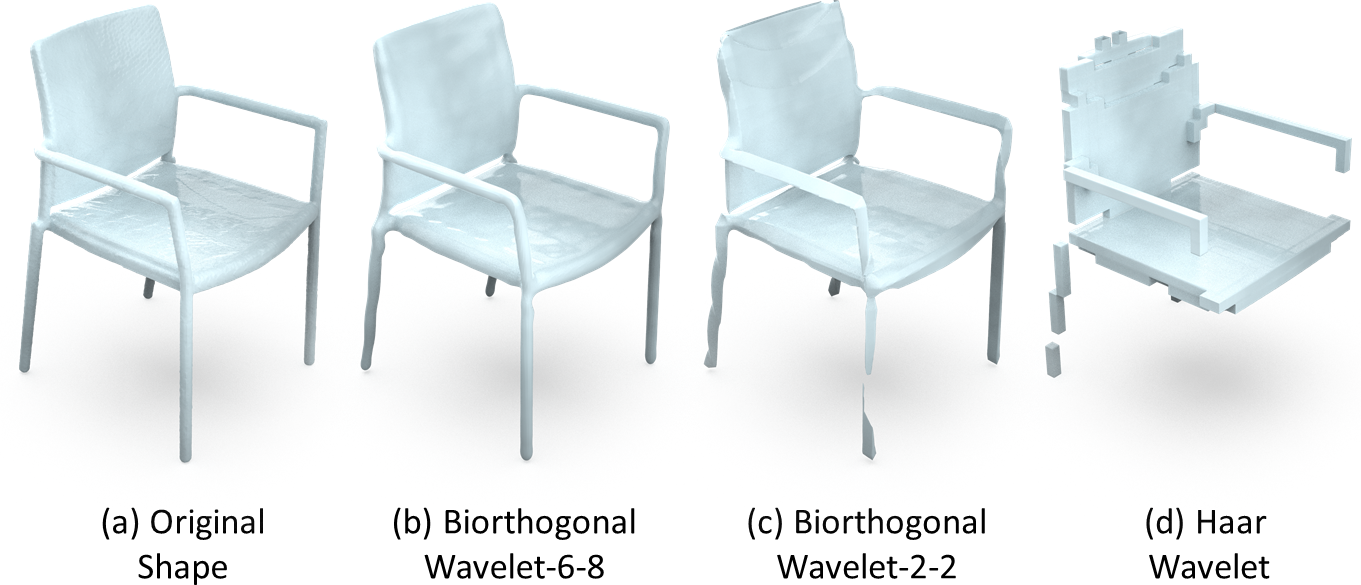}
	\vspace*{-1mm}
	\caption{Reconstructions with different wavelet filters.
	(a) An input shape from ShapeNet.
	(b,c) Reconstructions from the $J$=3 coefficient volumes with biorthogonal wavelets.
	The two numbers mean the vanishing moment of the synthesis and analysis wavelets.
	(d) Reconstruction with the Haar wavelet.}
	\label{fig:compact_analysis}
	\vspace*{-3mm}
\end{figure}

There are three important considerations in the data preparation.
First, multi-scale decomposition can effectively separate rich structures, fine details, and noise in the TSDF.
Empirically, we evaluate the reconstruction error on the TSDF by masking out all higher-scale detail coefficients and reconstructing $S_i$ only from the coefficients at scale $J=3$,~\ie, $C^3_i$ and $D^3_i$.
We found that the reconstructed TSDF values have relatively small changes from the originals (only 2.8\% in magnitude), even without 97\% of the coefficients
for the Chair category in ShapeNet~\cite{chang2015shapenet}.
Comparing Figures~\ref{fig:compact_analysis} (a) vs. (b), we can see that reconstructing only from the coarse scale of $J=3$ already well retains the chair's structure.
Motivated by this observation, we propose to construct the compact wavelet representation at a coarse scale ($J=3$)
and drop other detail coefficient volumes,~\ie, $D^1_i$ and $D^2_i$,
for efficient shape learning.
See supplementary material Section I for more details on the wavelet decomposition.

Second, we need 
a suitable wavelet filter.
While Haar wavelet is a popular choice due to its simplicity,
using it to encode smooth and continuous signals such as the SDF may introduce
serious voxelization artifacts, see,~\eg, Figure~\ref{fig:compact_analysis}(d).
In this work, we propose to adopt the biorthogonal wavelets~\cite{cohen1992biorthogonal}, since it
enables a more smooth decomposition of the TSDF.
Specifically, we tried different settings in the biorthogonal wavelets and chose to use high vanishing moments: six for the synthesis filter and eight for the analysis filter; see 
Figures~\ref{fig:compact_analysis}(b) vs. (c).
Also, instead of storing the detail coefficient volumes with seven channels, as in traditional wavelet decomposition, we follow~\cite{velho1994multiscale} to efficiently compute it as the difference between the inverse transformed $C^{j}_i$ and $C^{j-1}_i$ in a Laplacian pyramid style.
Hence, the detail coefficient volume has a higher resolution than the coarser one, but both have much lower resolution than the original TSDF volume ($256^3$).

Last, it is important to truncate the SDF before constructing the wavelet representation for shape learning.
By truncating the SDF, 
regions not close to the shape surface would be cast to 
a constant function to make efficient the wavelet decomposition and shape learning.
Otherwise, we found that 
the shape learning process will collapse and the training loss cannot be reduced.

\subsection{Shape Learning}
\label{ssec:shape_learning}

Next, to learn the 3D shape distribution in a given shape set, we gather coefficient volumes $\{C_i^J , D_i^J\}$ of the shapes in the set for training
(i) the {\em generator network\/} to learn to iteratively remove noise from a random Gaussian noise sample to generate $C_i^J$; and
(ii) the {\em detail predictor network\/} to learn to predict $D_i^J$ from $C_i^J$ to enhance the details in the generated shapes.
Further, to enable shape inversion, we may additionally train (iii) the {\em encoder network\/} (jointly optimized with the {\em generator network\/}) to learn a latent space for mapping the coarse coefficient volume $C_i^J$ to a latent code $z_{i}$.


\vspace*{-3pt}
\paragraph{Network structure} \
To start, we formulate a simple but efficient neural network structure for both the generator and detail predictor networks.
The two networks have the same structure, as both take a 3D volume as input and output a 3D volume of the same resolution as the input.
Specifically, we adopt a modified 3D version of the U-Net architecture~\cite{nichol2021improved}.
First, we apply three 3D 
residual blocks
to progressively compose and downsample the input into a set of multi-scale features and a bottleneck feature volume.
Then, we apply a single self-attention layer to aggregate features in the bottleneck volume, so that we can efficiently incorporate non-local information into the features.
Further,
we upsample and concatenate features in the same scale and progressively perform an inverse convolution 
with three residual blocks
to produce the output.
For all convolution layers in the network structure, we use a filter size of three with a stride of one.

For the encoder network, we design a five-layer 3D convolutional neural network with kernel size $k=4$ and stride $s=1$, each followed by an instance normalization.
Also, we adopt a single linear transform
to produce the output latent code $z$.


In the followings, we will first introduce the modeling of the shape generation process, followed by the adaptation for the shape inversion process.
Lastly, we will introduce the detail predictor network for enhancing the details of the generated shapes.

\paragraph{Modeling the shape generation process.} \
We formulate the 3D shape generation process based on the denoising diffusion probabilistic model~\cite{sohl2015deep}.
For simplicity, we drop the subscript and superscript in $C_i^J$ , and denote $\{ C_{0}, ..., C_{T} \}$ as the shape generation sequence, where
$C_0$ is the target, which is $C_i^J$;
$C_T$ is a random noise volume from the Gaussian prior; and
$T$ is the total number of time steps.
As shown on top of Figure~\ref{fig:overview}(b), we have 
(i) a forward process (denoted as $q(C_{0:T})$) that progressively adds noise based on a Gaussian distribution to corrupt $C_0$ into a random noise volume; and
(ii) a backward process (denoted as $p_{\theta}(C_{0:T})$) that employs the generator network (with network parameter $\theta$) to iteratively remove noise from $C_T$ to generate the target.
Note that all 3D shapes $\{ C_{0}, ..., C_{T} \}$ are represented as 3D volumes and each voxel value is a wavelet coefficient at its spatial location.

Both the forward and backward processes are modeled as the Markov processes.
The generator network is optimized to maximize the generation probability of the target, \ie, $p_{\theta}(C_0)$.
Also, as suggested in~\cite{ho2020denoising}, this training procedure can be further simplified
to use the generator network to predict the noise volume $\epsilon_{\theta}$.
So, we adopt 
a mean-squares loss to train our framework: 
\begin{equation}
    \label{eq:objective_simp}
    L_2 = E_{t,C_0,\epsilon}[{\parallel} \epsilon - \epsilon_{\theta}(C_t, t) {\parallel}^2], \epsilon \sim \mathcal{N}(0,\mathbf{I}),
\end{equation}
where 
$t$ is the time steps;
$\epsilon$ is a noise volume; and
$\mathcal{N}(0,\mathbf{I})$ denotes a unit Gaussian distribution.
In particular, we first sample noise volume $\epsilon$ from a unit Gaussian distribution $\mathcal{N}(0,\mathbf{I})$ and 
time step $t \in [1,...,T]$ to corrupt $C_0$ into $C_t$.
Then, our generator network learns to predict noise $\epsilon$ based on the corrupted coefficient volume $C_t$. 
Further, as the network takes 
time step
$t$ as input, we convert value $t$ into an embedding via two MLP layers.
Using this embedding, we can condition all the convolution modules in the prediction and enable the generator to be more aware of the amount of noise contaminated in $C_t$.
For more details on the derivation of the training objectives, please refer to supplementary material Section J.

\paragraph{Modeling the shape inversion process.}

The shape inversion process aims to 
embed a given 3D shape as a compact code $z$, such that we can faithfully reconstruct the coarse wavelet volume $C_0$ of the shape from $z$ and further reconstruct the shape.
Motivated by~\cite{preechakul2022diffusion}, we 
introduce an optional encoder $Enc_{\phi}$ to derive the latent code $z$ of coarse wavelet volume $C_0$,~\ie, $z = Enc_{\phi}(C_0)$.
Then, we inject $z$ into the generator during the backward process as an additional shape condition; see the bottom part of Figure~\ref{fig:overview}(b).
During the training, we jointly optimize the encoder and generator networks to maximize the conditional generation probability,~\ie, $p_\theta(C_0|z)$.
This can be achieved by modifying the objective in Equation~\eqref{eq:objective_simp} into
\begin{equation}
    \label{eq:objective_inv}
    L_2 = E_{t,C_0,\epsilon}[{\parallel} \epsilon - \epsilon_{\theta}(C_t, z, t) {\parallel}^2], \epsilon \sim \mathcal{N}(0,\mathbf{I}).
\end{equation}
Note that since we need the generator to be aware of the current shape condition, we follow~\cite{preechakul2022diffusion} to use the group normalization layers to jointly take the time embedding and the latent code as the input 
to the generator.

\vspace*{-3pt}

\paragraph{Detail predictor network}
Next, we train the detail predictor network to produce the detail coefficient volume $D_0$ from coarse coefficient volume $C_0$ (see the top part of Figure~\ref{fig:overview}(b)), so that we can further enhance the details in our generated (or inverted) shapes.

To train the detail predictor network, we leverage the paired coefficient volumes $\{ C_i^J, D_i^J \}$ from the data preparation.
Importantly, each detail coefficient volume $D_0$ should be highly correlated to its associated coarse coefficient volume $C_0$.
Hence, we pose detail prediction as a conditional regression on the detail coefficient volume, aiming at learning neural network function $f: C_0 \rightarrow D_0$; hence, we optimize $f$ via
a mean squared error loss.
Overall, the detail predictor has the same network structure as the generator, but we include more convolution layers to accommodate the cubic growth in the number of 
nonzero values
in the detail coefficient volume.

\begin{figure}[t]
	\centering
	\includegraphics[width=0.999\linewidth]{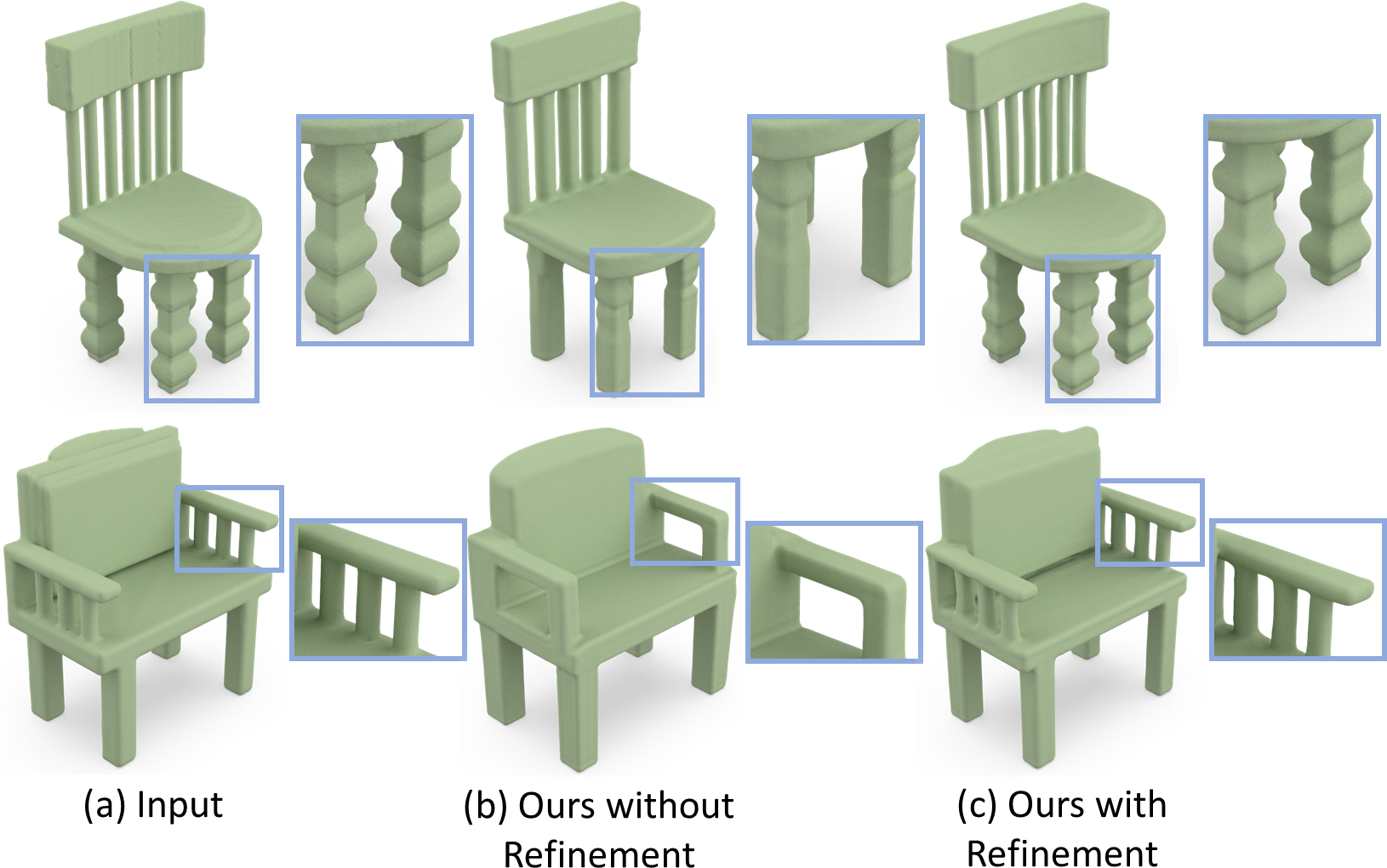}
	\vspace*{-5mm}
	\caption{Visual comparison of shape inversion with/without shape-guided refinement.
	Directly using our latent codes without refinement can already produce plausible shapes with overall appearance very similar to the inputs.
	Further refinement can enhance the reconstruction of the local structures,~\eg, see the chair legs on top and the chair armrests on bottom.
	}
	\label{fig:comp_op}
\end{figure}

\begin{figure*}[!t]
 	\centering
 	\includegraphics[width=1.0\linewidth]{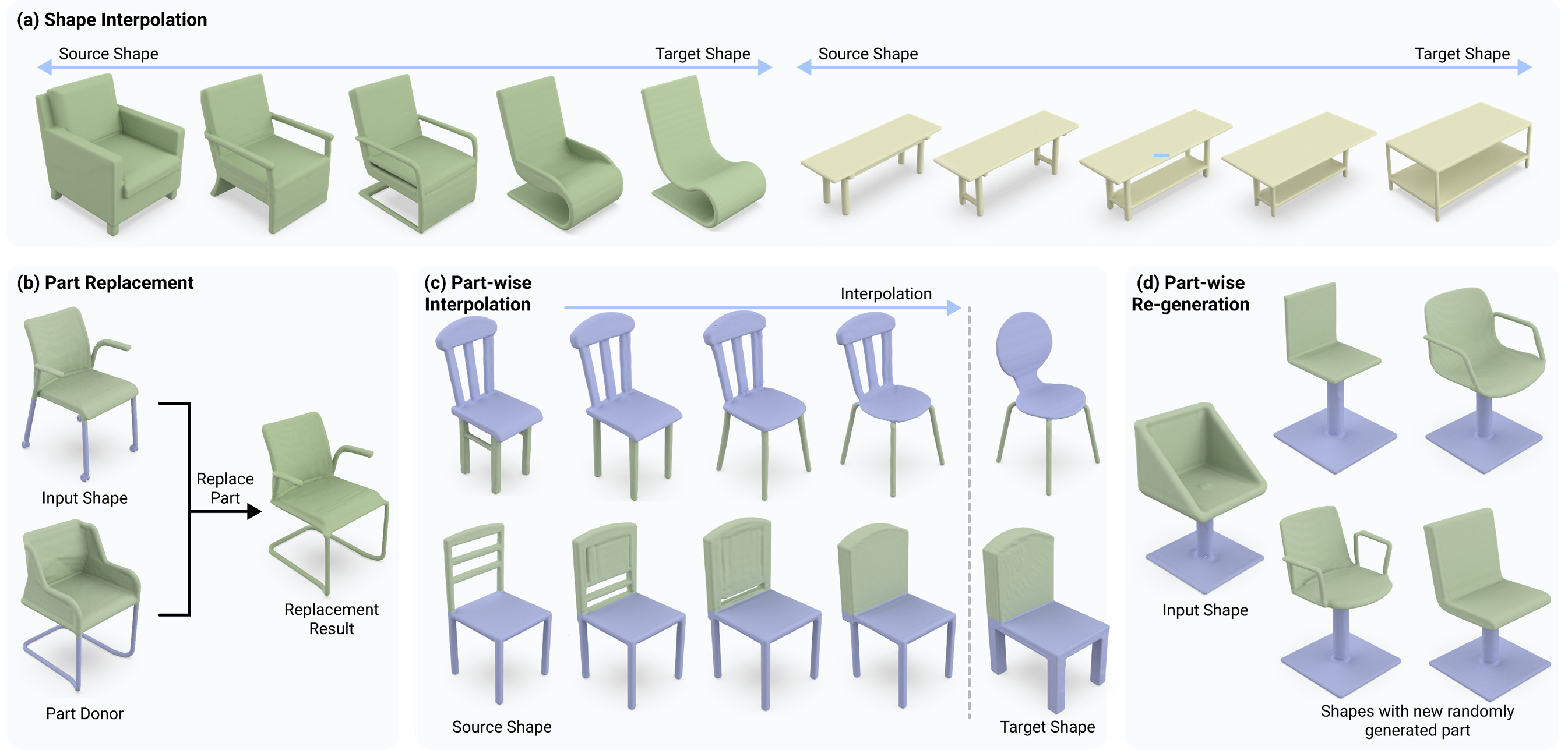}
 	\vspace*{-7mm}
 	\caption{
    Shape manipulations supported by our method.
    (a) Shape interpolation.
    We can smoothly interpolate inverted shapes to others using the learned latent space. 
    (b) Part replacement.
    We can replace a part in a shape (in blue) with a part from a donor shape (in blue).
    (c) Part-wise shape interpolation. 
    We can continuously interpolate
    only the selected part region (in green) in the source shape. 
    (d) Part-wise shape re-generation.
    We can re-generate a selected part (in green) in a shape while keeping the other parts untouched (in blue).
    The re-generated parts are diverse, plausible, and consistent with the untouched parts.
  }
 	\label{fig:manipulate_demo}
\end{figure*}

\subsection{Shape Generation}
\label{ssec:shape_generation}
Now, we are ready to generate 3D shapes.
Figure~\ref{fig:overview}(c) illustrates the shape generation procedure.
First, we randomize a 3D noise volume as $C_T$ from the standard Gaussian distribution.
Then, we can employ the trained generator 
for $T$ 
time steps
to produce $C_0$ from $C_T$.
This process is iterative and inter-dependent.
We cannot parallelize the operations in different 
time steps, so leading to 
a very long computing time.
To speed up the inference process,
we adopt an approach in~\cite{song2020denoising} to sub-sample a set of  time steps from $[1,..., T]$ during the inference; in practice, we evenly sample $1/10$ of the total time steps in all our experiments.

After we obtain the coarse coefficient volume $C_0$, we can then use the detail predictor network to predict the detail coefficient volume $D_0$ from $C_0$, followed by
a series of inverse wavelet transforms from $\{ C_0 , D_0 \}$ at scale $J$=$3$ to reconstruct the original TSDF.
By then, we can extract an explicit 3D mesh from the reconstructed TSDF using the marching cube algorithm~\cite{lorensen1987marching}.

\subsection{Shape Inversion}
\label{sec:inversion}
Thanks to the encoder network trained during the shape learning, our approach can leverage it for shape inversion.
Having said that, our goal is to invert a given unseen shape into a latent code and then reconstruct it faithfully from the latent code.

Figure~\ref{fig:overview}(d) illustrates the overall shape inversion procedure.
First, we produce coarse coefficient volume $C_0$ from the input shape, following the procedure in Section~\ref{sec:data_preparation}.
Then, we feed coefficient volume $C_0$ into the encoder network to obtain the latent code $z=Enc_{\phi}(C_0)$.
After that, we feed latent code $z$ together with a sampled noise volume $\epsilon$ into the generator network to directly produce a coarse coefficient volume for 
reproducing
the original shape, following the procedure in Section~\ref{ssec:shape_generation}.

However, 
directly generating the shape from latent code $z$ would lead to loss in topological structures and fine geometric details originally in the shape; compare Figure~\ref{fig:comp_op}(a) and (b).
To enhance the quality of the inverted shape, we further propose a 
shape-guided refinement scheme to
search for latent code $z'$ around $z$ in the latent space to better fit the input shape $C_0$.
In detail, we initialize latent code $z'$ as $z$ and 
adapt it by gradient descent using the inversion objective in Equation~\eqref{eq:objective_inv} (see Section~\ref{ssec:shape_learning}) for 400 iterations
on the input.
Using the initial latent code $z$, we can already obtain a plausible shape similar to the input.
By using this shape-guided refinement, we can further obtain fine details and structures in local regions missed in the initial code $z$.
Also, note that during the refinement, both the encoder and generator networks are fixed.
As Figure~\ref{fig:comp_op}(c) shows, 
our refined latent code helps encourage more faithful shape inversion with precise topological structures and fine geometric details, more similar to the original inputs.
In Section~\ref{sec:result_inversion}, we will present quantitative evaluation of our results, showing that our approach can produce inverted shapes of much higher fidelity, compared with the state-of-the-art methods.


\subsection{Shape Manipulation}
\label{sec:manipulation}

Shape inversion enables us to faithfully encode shapes into the learned latent space, in which our latent codes can faithfully represent their associated 3D shapes.
Further, we design the shape generation process with the latent code as a condition.
Hence, by manipulating the latent code and regenerating shape from the manipulated code, we can produce new shapes from the existing ones, e.g., 
by simply interpolating latent codes of different shapes; see our high-quality shape interpolation results shown in Figure~\ref{fig:manipulate_demo}(a).

Further than that, our method enables a rich variety of region-aware manipulations,
in which we can manipulate a particular region of a shape, while leaving other regions untouched:
\begin{itemize}
%
%

\item[(i)] \para{Part Replacement.} \
First, we can 
replace a selected part in the input shape with a part from the donor;
see Figure~\ref{fig:manipulate_demo}(b) for an example.
Note that the new shape is still plausible and the new part is consistent with the other parts.
%
%
\item[(ii)] \para{Part-wise Interpolation.} \
Instead of interpolating the whole shape, our method allows us to select a part in the input and interpolate the part region towards another shape; see Figure~\ref{fig:manipulate_demo}(c).
We can observe that changes in the selected region are smooth and semantically meaningful during the interpolation with good coherency across different parts.
%
%
%
\item[(iii)] \para{Part-wise Re-generation}.
Further, we can randomly re-generate a selected part in the input; see the upper part of the chair in Figure~\ref{fig:manipulate_demo}(d).
Our re-generated parts are diverse, plausible, and consistent with the untouched parts.
\end{itemize}

\begin{figure}[t]
	\centering
	\includegraphics[width=0.95\linewidth]{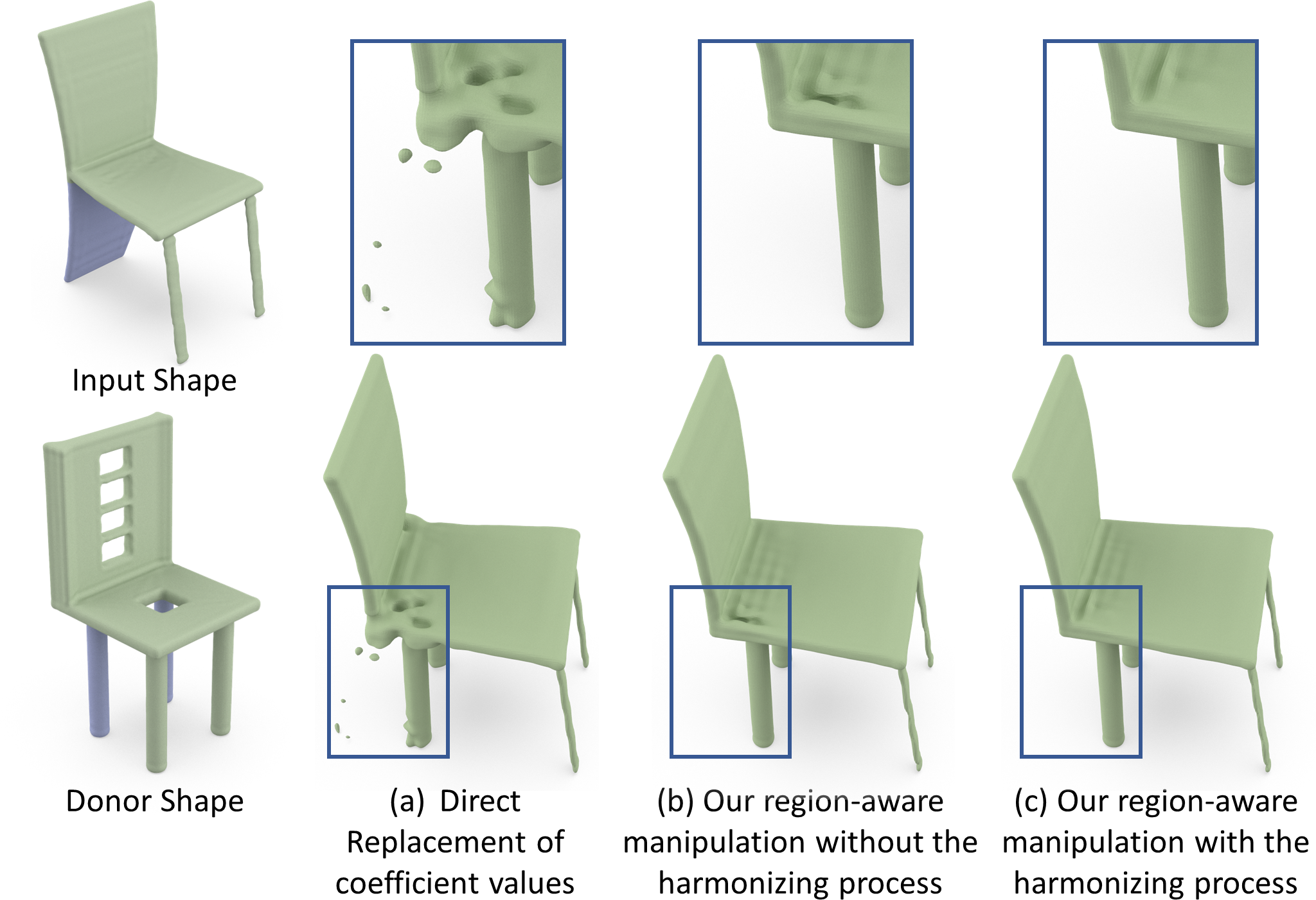}
	\vspace{-1mm}
	\caption{
    Visual comparisons on part replacement by:
    (a) a direct replacement of coefficient values;
    (b) our region-aware manipulation without the harmonizing process;
    and (c) our full region-aware manipulation procedure with the harmonizing process.
        Directly replacing the coefficient values leads to noise near the boundaries between the new part from the donor shape and the remaining parts in the input shape; see (a).
        Without the harmonizing process, inconsistencies in the coefficient volume could introduce artifacts on the manipulated shape; see (b).
        With our full approach, we can smooth out the connected regions for more consistent part replacement; see (c).
    }
	\label{fig:ablation_harmonize}
	\vspace{-2mm}
\end{figure}

To support these region-aware shape manipulations, we utilize the property that 
our generated coarse coefficient volume $C_0$ can maintain a good spatial correspondence with the original TSDF volume.
Hence, one can select a region in a shape and locally manipulate the associated portion in the coarse coefficient volume.
However, naively changing the coefficient values 
in the selected region will introduce noise around the connecting boundary, since the new coefficient values may not be consistent with the original coefficients in the other regions of the shape;
see,~\eg, Figure~\ref{fig:ablation_harmonize}(a).

\begin{figure*}[!t]
 	\centering
 	\includegraphics[width=1.0\linewidth]{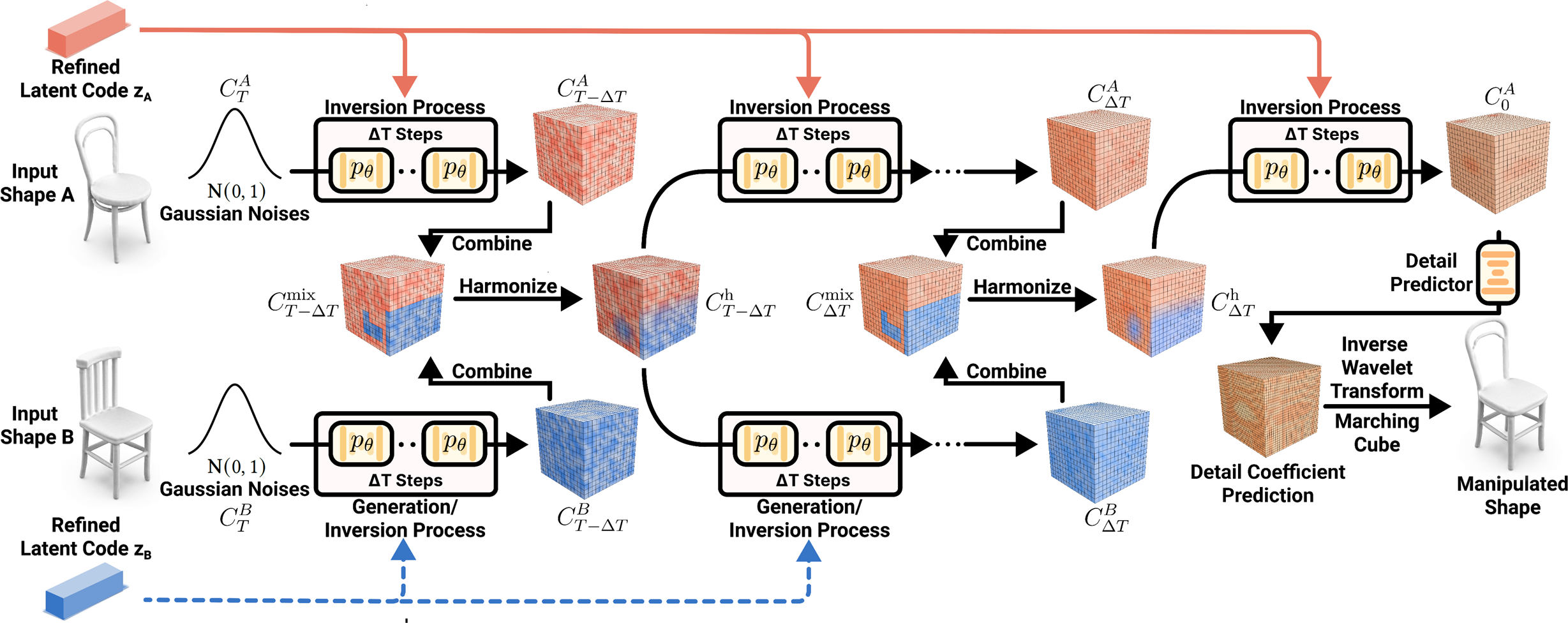}
 	\vspace*{-5mm}
 	\caption{Overview of our region-aware manipulation procedure, taking $T$ steps to produce the manipulated coarse coefficient volume $C^{A}_{0}$ from the sampled 3D Gaussian noise volume $C^{A}_{T}$.
 	First, we run two inversion processes (left side) for $\Delta T$ steps in parallel, guided by the two refined latent codes $z_A$ and $z_B$ from input shapes A and B to produce two partially-denoised coefficient volumes $C^A_{T-\Delta T}$ and $C^B_{T-\Delta T}$, respectively.
 	We then spatially combine the coefficient values of two coefficient volumes to obtain the mixed coefficient volume $C^{\text{mix}}_{T-\Delta T}$ and further harmonize values in the boundary regions to produce $C^{\text{h}}_{T-\Delta T}$.
 	With the harmonized coefficient volume as guidance,
  we repeat this combine-and-harmonize process every $\Delta T$ steps until we produce the manipulated coefficient volume $C^{A}_{0}$. 
 	Also, we can replace the bottom inversion process with an unconditional shape generation process for achieving part-wise re-generation. 
 	}
	\label{fig:overview_manipulate}
\end{figure*}

To address this issue, we propose the region-aware manipulation procedure shown in Figure~\ref{fig:overview_manipulate} by extending the shape 
inversion
pipeline.
Overall, the diffusion process takes $T$ steps to produce the manipulated coarse coefficient volume $C^{A}_{0}$ from the random noise volume $C^{A}_T$.
See Figure~\ref{fig:overview_manipulate} (top left), given input shape A and its refined latent code $z_A$ (from Section~\ref{sec:inversion}), we first run the shape inversion process for $\Delta T$ steps to obtain the partially-denoised coefficient volume $C^A_{T-\Delta T}$.
In parallel, see Figure~\ref{fig:overview_manipulate} (bottom left), we do the same on input shape B (which can be the donor/target shape, depending on the type of the manipulation operation) and its refined code $z_B$
to produce another coefficient volume $C^{B}_{T-\Delta T}$.

Importantly, after every $\Delta T$ steps in the diffusion process (where $T = M \cdot \Delta T$ for some positive integer $M$), 
we replace the coefficient values in the selected region of $C^{A}_{T-\Delta T}$ by those in $C^{B}_{T-\Delta T}$ to produce the mixed coefficient volume $C^{\text{mix}}_{T-\Delta T}$.
As the coefficient values near the mixed boundary in $C^{\text{mix}}_{T-\Delta T}$ may not be consistent from $C^{A}_{T-\Delta T}$ and $C^{B}_{T-\Delta T}$, the generated shapes may contain some small artifacts; see,~\eg, Figure~\ref{fig:ablation_harmonize}(b).
To smooth the coefficient mixing, we propose to adopt the harmonizing process in~\cite{lugmayr2022repaint}
to produce the harmonized coefficient volume $C^{\text{h}}_{T-\Delta T}$ (details in the next paragraph).
By harmonizing $C^{\text{mix}}_{T-\Delta T}$ after combining $C^A_{T-\Delta T}$ and $C^B_{T-\Delta T}$, we can obtain a smooth transition of coefficient values near the boundary; see the improved result in Figure~\ref{fig:ablation_harmonize}(c).
After that, the harmonized coefficient volume can guide the subsequent steps of the two processes, so we use a combine-and-harmonize process in every $\Delta T$ steps to obtain the final manipulated coefficient volume; see again Figure~\ref{fig:overview_manipulate}.
We empirically set $\Delta T = 10$ for region-aware manipulation experiments.
Also, we can replace the inversion process guided by the refined latent code $z_B$ with an unconditional shape generation for achieving part-wise re-generation.
For the details on how we select the manipulation region in the input shape and how we compute the corresponding region in the wavelet domain; please refer to supplementary material Section F.

\begin{figure*}[t]
	\centering
	\includegraphics[width=0.97\linewidth]{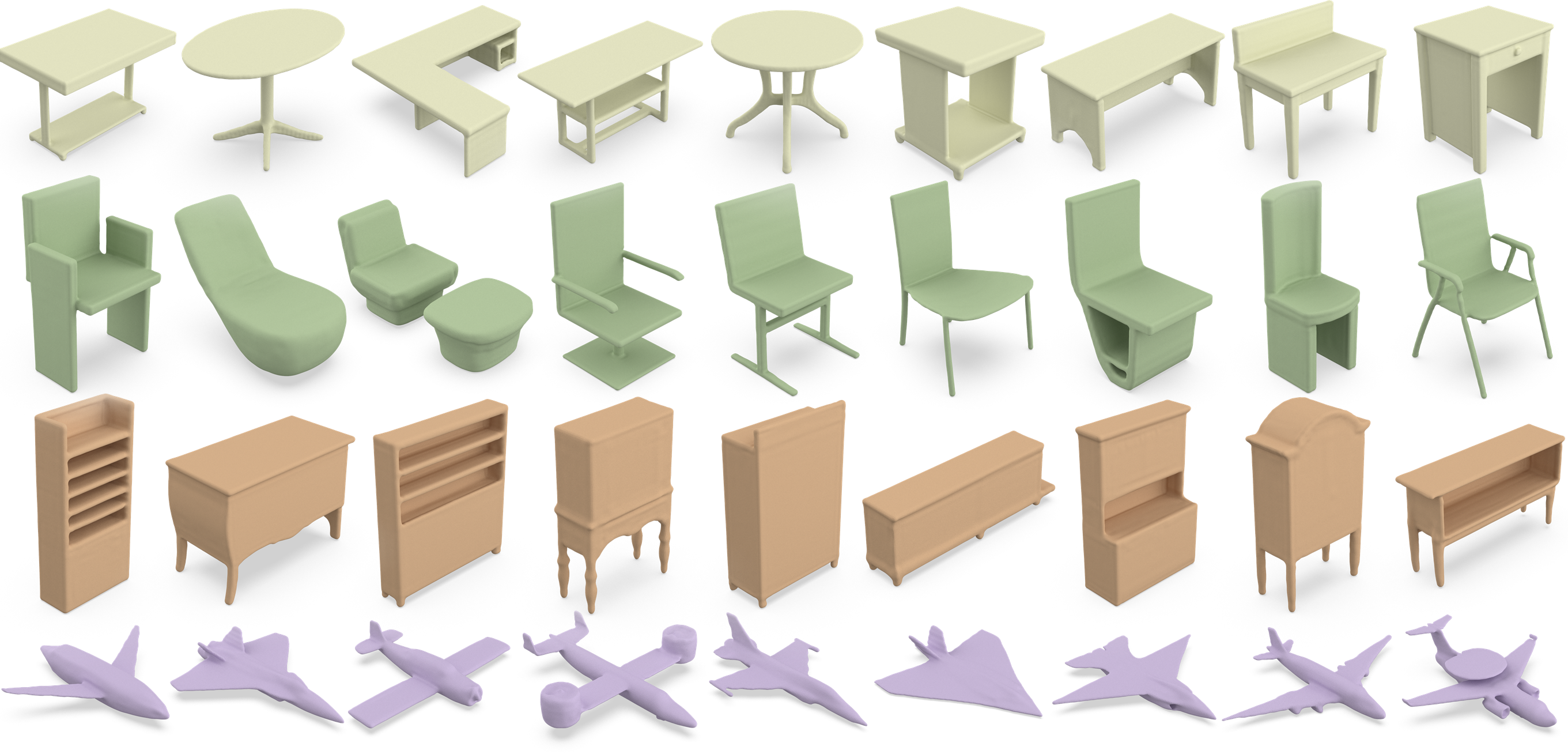}
	\vspace*{-4.5mm}
	\caption{
	Gallery of our generated shapes: Table, Chair, Cabinet, and Airplane (top to bottom).
	Our shapes exhibit complex structures, fine details, and clean surfaces, without obvious artifacts, compared with those generated by the other approaches; see Figure~\ref{fig:query_comapre}.}
	\label{fig:gallery}
	\vspace*{-3mm}
\end{figure*}

\begin{figure}[t]
	\centering
	\includegraphics[width=0.99\linewidth]{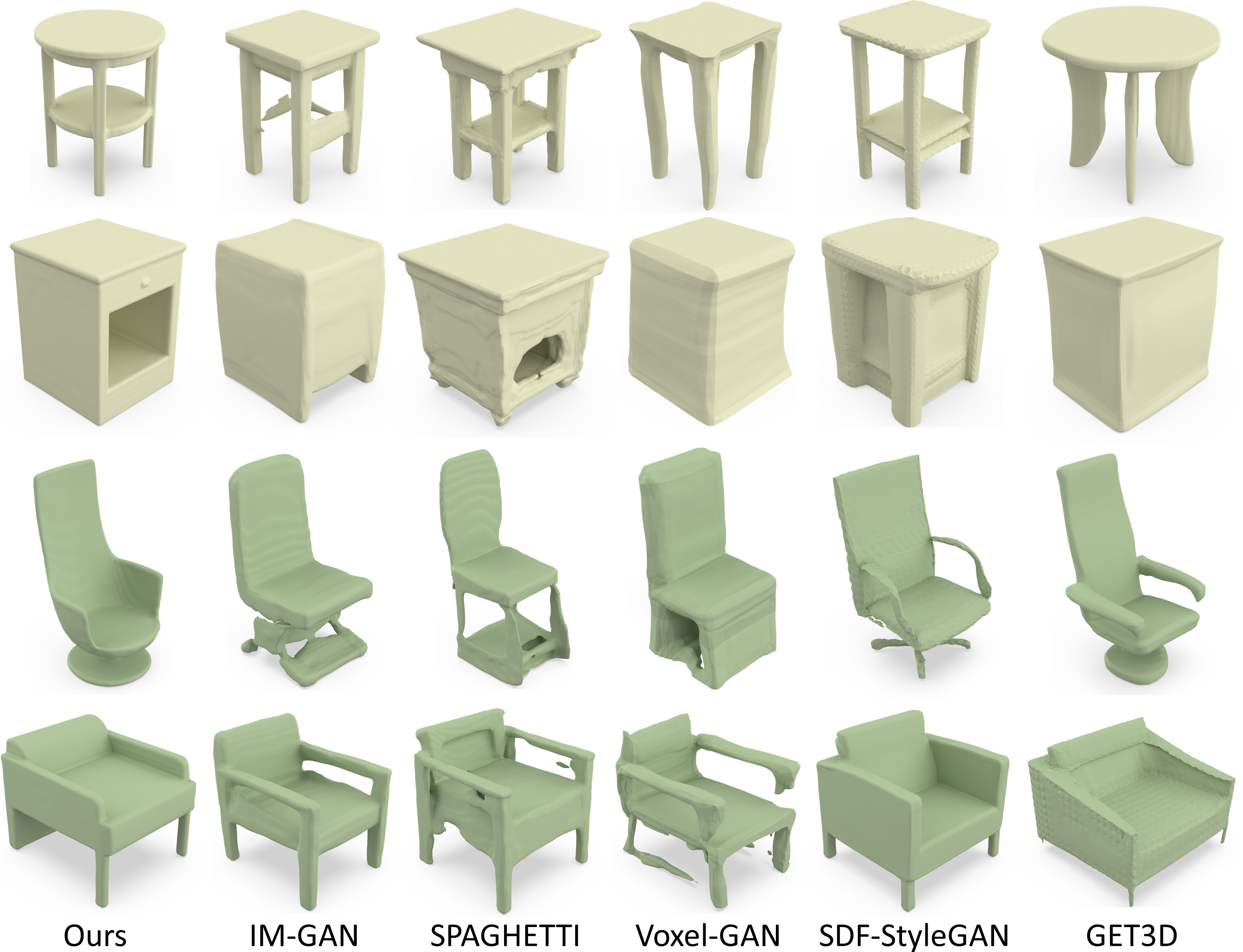}
	\vspace{-3mm}
	\caption{Visual comparisons with state-of-the-art methods.
	Our generated shapes exhibit finer details 
 and cleaner surfaces, without obvious artifacts.
	}
	\label{fig:query_comapre}
	\vspace{-3mm}
\end{figure}
\paragraph{Details on the harmonizing process}
Given the mixed 
coefficient volume $C^{\text{mix}}_t$, we follow the forward process of the diffusion model to add noise to it by sampling $C^{\text{mix}}_{t+1} \sim \mathcal{N}(\sqrt{1 - \beta_t} {C}_{t}, \beta_t \mathbf{I})$.
Then, we apply the two inversion processes mentioned above, guided by the refined latent codes $z_A$ and $z_B$, respectively,
separately on $C^{\text{mix}}_{t+1}$,
for a single step to obtain two denoised coefficient volumes.
We then combine the volumes according to the selected region again and repeat the above adding noise and denoising procedure ten times to obtain the harmonized coefficient volume.
By doing so, the generator can better account for the coefficient value changes in the manipulated region and adapt the values near the boundary.
In the case of part-wise re-generation, we use the unconditional shape generation for harmonizing the mixed coefficient volume instead of the inversion process guided by the refined latent code $z_B$.

\subsection{Implementation Details}
\label{sec:implementation}

We employed ShapeNet~\cite{chang2015shapenet} to prepare the training dataset used in all our experiments.
For shape generation, we follow the data split in~\cite{chen2019learning} and use only the training split to supervise our network training.
For shape inversion, we follow the data split in ~\cite{park2019deepsdf} for ease of comparison.
Also, similar to~\cite{hertz2022spaghetti,luo2021diffusion,li2021spgan}, we train one model of each category in the ShapeNet dataset~\cite{chang2015shapenet} for both shape generation and inversion.

We implement our networks using 
PyTorch and run all experiments on a GPU cluster with four RTX3090 GPUs.
For both shape generation and inversion,
 we follow~\cite{ho2020denoising} to set $\{\beta_t\}$ to increase linearly from $1e^{-4}$ to $0.02$ for 1,000 steps and 
set $\sigma_t = \frac{1-\bar{\alpha}_{t-1}}{1 - \bar{\alpha}_t} \beta_t$.
We train the generator, optionally with the encoder, for 800,000 iterations and
the detail predictor for 60,000 iterations, both 
using the Adam optimizer~\cite{kingma2014adam} with a learning rate of $1e^{-4}$.
Training the generator and detail predictor takes around three days and 12 hours, respectively.
For shape-guided refinement, we also adopt Adam optimizer~\cite{kingma2014adam} with a learning rate of $5e^{-2}$ for 400 iterations.
For shape generation, the inference takes around six seconds per shape on an RTX 3090 GPU.
For shape inversion, the refinement procedure and inference totally take around two minutes on an RTX 3090 GPU using 1000 diffusion steps.
As for shape manipulation, our region-aware manipulation procedure takes four minutes on an RTX 3090 GPU for running 1000 diffusion steps.
We adapt~\cite{cotter2020uses} to implement the 3D wavelet decomposition.
{\em We will release our code and training data upon the publication of this work.\/}
\input{tables/quantitative.tex}

%% file: tables/quantitative.tex
\begin{table*}[t]
	\centering
		\caption{Quantitative comparison between the generated shapes produced by our method and six state-of-the-art methods.
		We follow the same setting to conduct this experiment as in the state-of-the-art methods.
		From the table, we can see that our generated shapes have the best quality for almost all cases
		(largest COV, lowest MMD, and 1-NNA close to 50) for both the Chair and Airplane categories.
The units of CD and EMD are $10^{-3}$ and $10^{-2}$, respectively.}
        \vspace*{-2mm}
	\resizebox{0.88\linewidth}{!}{
		\begin{tabular}{C{5cm}|C{0.6cm}C{0.6cm}C{0.6cm}C{0.6cm}C{0.6cm}C{0.6cm}|C{0.6cm}C{0.6cm}C{0.6cm}C{0.6cm}C{0.6cm}C{0.6cm}}
			\toprule[1pt]
                        \multirow{3}*{Method} & \multicolumn{6}{c|}{Chair}                                                  & \multicolumn{6}{c}{Airplane}
                        \\
                         & \multicolumn{2}{c}{COV$~\uparrow$} & \multicolumn{2}{c}{MMD$~\downarrow$} & \multicolumn{2}{c|}{1-NNA $\sim\hspace{-1mm}50$} & \multicolumn{2}{c}{COV$~\uparrow$} & \multicolumn{2}{c}{MMD$~\downarrow$} & \multicolumn{2}{c}{1-NNA $\sim\hspace{-1mm}50$} \\ 
                         & CD         & EMD        & CD         & EMD        & CD         & EMD        & CD         & EMD        & CD         & EMD        & CD         & EMD        \\ \hline
IM-GAN~\cite{chen2019learning}
& 56.49           & 54.50 & 11.79          & 14.52          & 61.98          & 63.45          & 61.55          & 62.79          & 3.320          & 8.371          & 76.21          & 76.08
\\ \hline
Voxel-GAN~\cite{kleineberg2020adversarial}
& 43.95           & 39.45 & 15.18          & 17.32          & 80.27          & 81.16          & 38.44          & 39.18          & 5.937          & 11.69          & 93.14          & 92.77
\\ \hline
Point-Diff~\cite{luo2021diffusion}
& 51.47           & 55.97 & 12.79          & 16.12          & 61.76          & 63.72          & 60.19          & 62.30          & 3.543          & 9.519          & 74.60          & 72.31
\\ \hline
SPAGHETTI~\cite{hertz2022spaghetti}
& 49.19           & 51.92 & 14.90          & 15.90          & 70.72          & 68.95          & 58.34          & 58.38          & 4.062          & 8.887          & 78.24          & 77.01          \\ \hline
SDF-StyleGAN~\cite{zheng2022sdfstylegan}
& 51.77           & 50.30 & 13.45          & 15.43          & 68.88          & 70.20          & 57.97          & 48.33          & 3.859          & 9.406          & 83.37          & 84.36          \\ \hline
GET3D~\cite{gao2022get3d}
& 53.47           & \textbf{56.41} & 14.43          & 15.63          & 70.32          & 69.51          & 55.62          & 55.38          & 4.134          & 9.421          & 89.12          & 86.77          \\ \hline \hline
Ours                     & \textbf{58.19}  & 55.46 & \textbf{11.70} & \textbf{14.31} & \textbf{61.47} & \textbf{61.62} & \textbf{64.78} & \textbf{64.40} & \textbf{3.230} & \textbf{7.756} & \textbf{71.69} & \textbf{66.74} \\
			\bottomrule[1pt]
	\end{tabular}
 }
    \vspace*{-1mm}
\label{tab:quanComparison}
\end{table*}

%% file: evaluation.tex
\vspace*{-1mm}
\section{Results and Experiments}

\subsection{Shape Generation}
\label{subsec:shape_gen}
\paragraph{Galleries of our generated shapes}

We present Figure~\ref{fig:teaser} (left) and Figure~\ref{fig:gallery} to showcase the compelling 
capability of our method for generating shapes of various categories.
Our generated shapes exhibit {\em diverse topology\/}, {\em fine details\/}, and also {\em clean surfaces without obvious artifacts\/}, covering a rich variety of small, thin, and complex structures that are typically very challenging for the existing approaches to produce.
More 3D shape generation results produced by our method are provided in supplementary material Section A. 

\vspace{-3pt}
\paragraph{Baselines for comparison} 
We compare the shape generation capability of our method
with six state-of-the-art methods:
IM-GAN~\cite{chen2019learning}, 
Voxel-GAN~\cite{kleineberg2020adversarial}, 
Point-Diff~\cite{luo2021diffusion}, 
SPAGHETTI~\cite{hertz2022spaghetti},
SDF-StyleGAN~\cite{zheng2022sdfstylegan},
and GET3D~\cite{gao2022get3d}.
To our best knowledge, our method is the first work that generates implicit shape representations in 
frequency domain and 
considers coarse and detail coefficients to enhance the generation of structures and fine details.

Our experiments follow the same setting as the above works.
Specifically, we leverage our trained model on the Chair and Airplane categories in ShapeNet~\cite{chang2015shapenet} to randomly generate 2,000 shapes for each category.
Then, we uniformly sample 2,048 points on each generated shape and evaluate the shapes using the same set of metrics as in the previous methods (details to be presented later).
For GET3D, we use the official pre-trained model for the Chair category and adopt their code to train a model for the Airplane category, which is not provided in their released repository.
For the other comparison methods, we employ publicly-released trained network models to generate shapes.

\vspace{-3pt}
\paragraph{Evaluation metrics for shape generation}
Following~\cite{luo2021diffusion,hertz2022spaghetti}, we evaluate the generation quality using
(i) minimum matching distance (MMD) measures the fidelity of the generated shapes;
(ii) coverage (COV) indicates how well the generated shapes cover the given 3D repository; and
(iii) 1-NN classifier accuracy (1-NNA) measures how well a classifier differentiates the generated shapes from those in the repository.
Overall, a low MMD, a high COV, and an 1-NNA close to 50\% indicate good generation quality;
see supplementary material Section K for the details.

\vspace{-3pt}
\paragraph{Quantitative evaluation for shape generation}
Table~\ref{tab:quanComparison} reports the quantitative comparison results,
showing that our method surpasses all others for almost all the evaluation cases over the three metrics for both the Chair and Airplane categories.
We employ the Chair category, due to its large variations in structure and topology, and the Airplane category, due to the fine details in its shapes.
As discussed in~\cite{yang2019pointflow,luo2021diffusion}, the COV and MMD metrics have limited capabilities to account for details, so they are not suitable for measuring the fine quality of the generation results,~\eg, the generated shapes sometimes show a better performance even when compared with the ground-truth training shapes on these metrics.
In contrast, 1-NNA is more robust and can better correlate with the generation quality.
In this metric, our approach outperforms all others, while having a significant margin in the Airplane category, manifesting the diversity and fidelity of our generated results.

\vspace{-3pt}
\paragraph{Qualitative evaluation for shape generation}
Figure~\ref{fig:query_comapre} shows some visual comparisons.
For each random shape generated by our method, we find a similar shape (with similar structures and topology) generated by each of the other methods to make the visual comparison easier.
See supplementary material Sections B and D for more visual comparisons.
Further, as different methods likely
have different statistical modes in the generation distribution, we also 
take random shapes generated by IM-GAN and find similar shapes generated by our method for comparison; see supplementary material Section~C for the results.
From all these results, we can see that the 3D shapes generated by our method clearly exhibit finer details, higher fidelity structures, and cleaner surfaces, without obvious artifacts.

\begin{figure*}[t]
	\centering
	\includegraphics[width=0.92\linewidth]{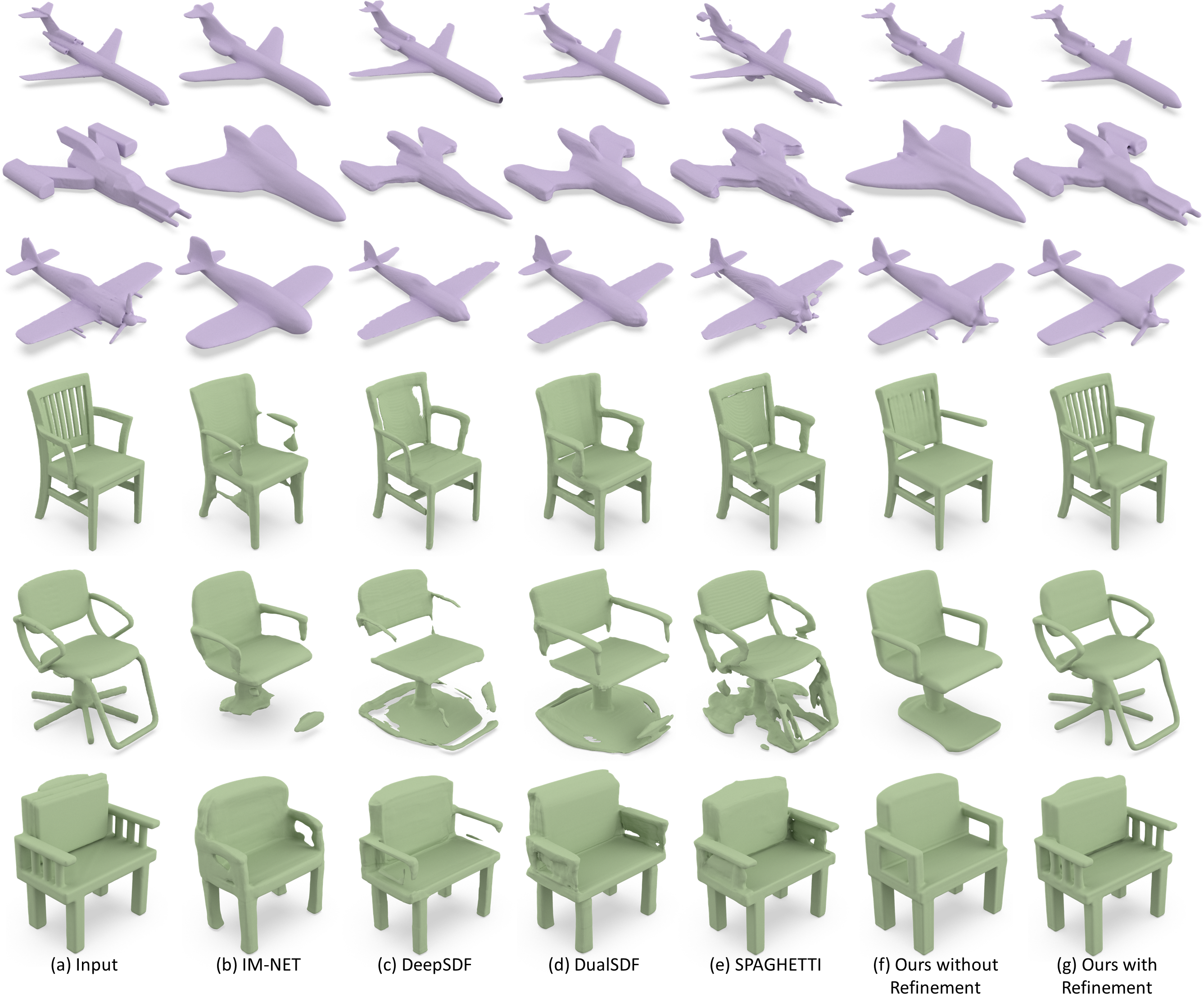}
	\vspace{-2mm}
	\caption{Visual comparisons on shape inversion.
	Our method is able to produce more faithful inverted shapes (g) that are highly similar to the inputs (a), compared with others (b-e). Our inverted results exhibit fine details and complete structures; see, \eg, the chair pulleys and airplane propellers, which 
	are typically very challenging for 
	the existing methods (b-e).
 Also, our method, without further refinements (f), can still produce reasonable shape inversions.
	}

	\label{fig:comp_inver}
	\vspace{-2mm}
\end{figure*}

\begin{figure}[t]
	\centering
	\includegraphics[width=0.95\linewidth]{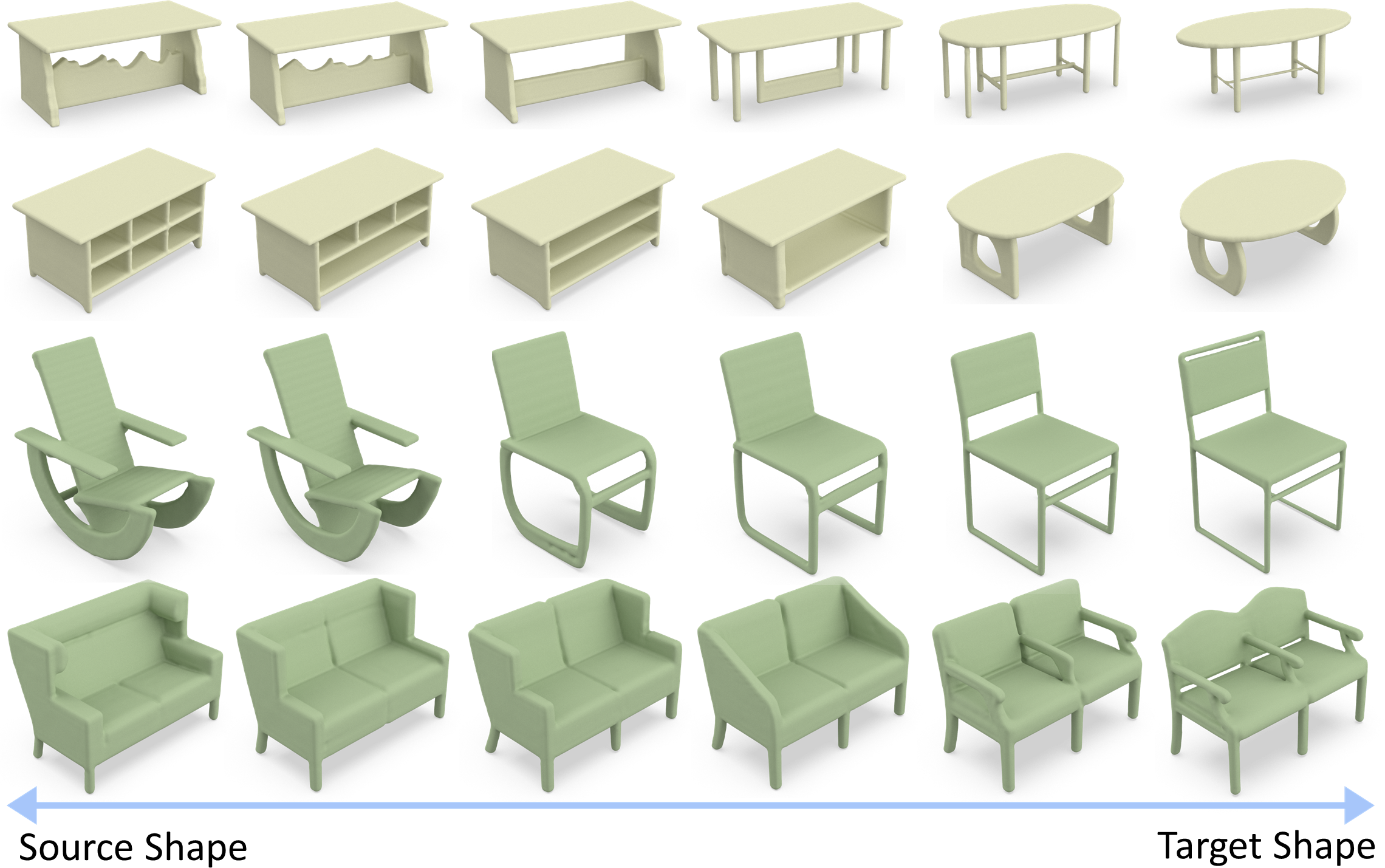}
	\vspace{-2mm}
	\caption{High-quality shapes created by interpolating our refined latent codes.
Note the smooth transition from the sources to the targets.
Particularly, the intermediate shapes are mostly plausible with fine details.}
	\label{fig:interpolation}
	\vspace{-2mm}
\end{figure}


\vspace{-3pt}
\subsection{Shape Inversion}
\label{sec:result_inversion}

\paragraph{Baselines for comparison}
We compare our shape inversion results with those from four state-of-the-art methods:
IM-NET~\cite{chen2019learning}, DeepSDF~\cite{park2019deepsdf}, DualSDF~\cite{hao2020dualsdf}, and SPAGHETTI~\cite{hertz2022spaghetti}.
We employ their official code to train their models, following the same train-test split as~\cite{park2019deepsdf} for a 
fair comparison.
Note that we do not evaluate the results of SPAGHETTI in the Lamp category, as its official pre-trained model is unavailable and the training code has not been officially released by the time of the submission.
Further, we notice a very recent work, NeuForm~\cite{lin2022neuform}, which
overfits the given shape for achieving the shape inversion.
Even though the code and pre-trained models of this work have not been released, we provide a visual comparison on the inversion results in 
supplementary material Section E using the example results given in their paper.

\vspace{-3pt}
\paragraph{Evaluation metrics for shape inversion}
Following prior works, we evaluate the inversion quality by measuring the similarity between the inverted shapes and the original inputs
using Chamfer Distance (CD), Earth Mover's Distance (EMD), and Light Field Distance (LFD)~\cite{chen2003visual}.
For CD and EMD, they evaluate point-wise distances between two point clouds sampled on the shape surfaces.
Here, we uniformly sample 2048 points on each shape (inverted and original) and evaluate the metrics on the sampled point clouds.
For LFD, it measures the difference between two shapes in the rendered image domain.
First, we uniformly sample 20 viewpoints to render images for each of the inverted shapes and input shapes.
Then, we compute a 45-dimension feature vector for each rendered image and obtain the final metric by summing up all pairwise L1 distances between the image feature vectors from the same viewpoint.
Note that a low value indicates a better performance for all three metrics.
Also, there could be noise in the sampled TSDF grid, so we 
post-process
the inverted shapes.
For the details on the metrics and 
post-processing,
 please see supplementary material Section L.

\input{tables/inversion_quantitative.tex}

\vspace{-3pt}
\paragraph{Quantitative evaluation for shape inversion}
Table~\ref{tab:inv_quanComparison} shows the quantitative comparison results.
Without refinement, our method already performs better in all metrics than IM-NET~\cite{chen2019learning}, which also does not require additional refinement.
By further refining the latent code,
our method can 
significantly outperform all the other methods in all metrics.
Particularly, for the LFD metric, our method manifests more than $30\%$ 
performance gain over the 
second-best
method in all categories.

\vspace{-3pt}
\paragraph{Qualitative evaluation for shape inversion}
Figure~\ref{fig:comp_inver} shows some visual comparisons on shape inversion.
We can observe that our method can already produce plausible results 
without refinement; see Figure~\ref{fig:comp_inver}(f).
By further introducing the shape-guided refinement, we can faithfully reproduce the fine details and complex structures in the input shapes; see,~\eg, the thin structures at the chair's back and the aircraft's propeller shown in the third and fourth rows of Figure~\ref{fig:comp_inver}; they cannot be achieved by the existing works.

\begin{figure}[t]
	\centering
	\includegraphics[width=0.95\linewidth]{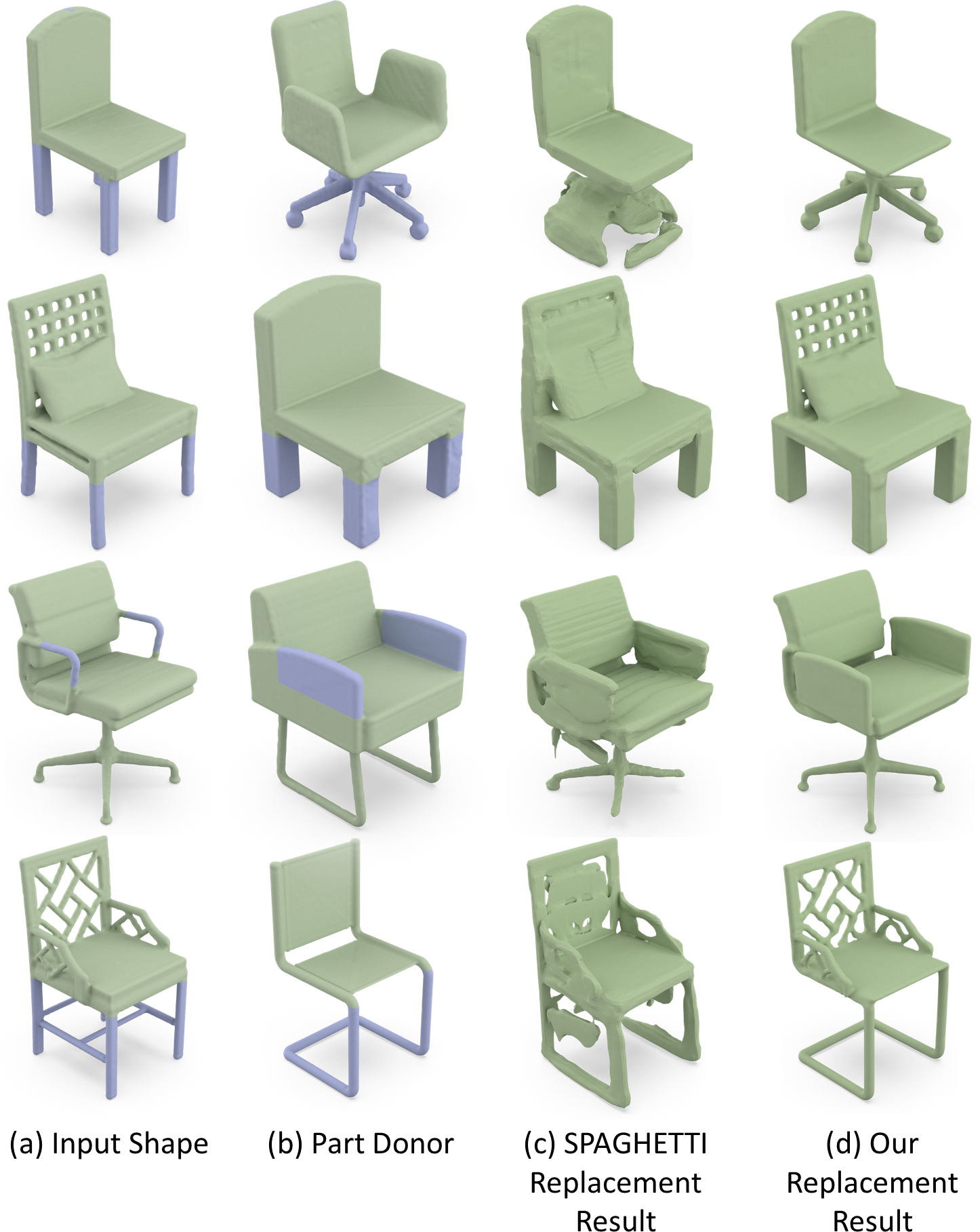}
	\vspace{-2mm}
	\caption{Part replacement results.
 We replace the blue part in the original shape (a) with the 
 blue part in the donor shape (b), thus generating new shapes (d).
 Compared with SPAGHETTI (c), our method can better preserve the original parts,~\eg, legs and castors, and produce plausible shapes.}
	\label{fig:shape_mixing}
	\vspace{-2mm}
\end{figure}

\vspace{-3pt}
\subsection{Shape Manipulation}
\label{sec:result_manipulation}

\paragraph{Shape Interpolation.}
As Figure~\ref{fig:interpolation} shows, our method can produce smooth and plausible interpolations between two unseen shapes, thanks to 
the shape-guided refinement.
From left to right, the source can morph smoothly towards the target;
see especially the consistent changes in the armrests in the last row of Figure~\ref{fig:interpolation}.
These results manifest the superior capability of our framework
to embed an unseen shape in a smooth and plausible latent space.

\vspace{-3pt}
\paragraph{Part Replacement.}
Besides, we can select a part in an input shape and replace it with a corresponding part in the donor shape.
As Figure~\ref{fig:shape_mixing} shows,
we replace the blue part in each input (a) with the blue part in the corresponding donor (b) to generate new shapes (d).
Note that the new shapes exhibit high fidelity,
while preserving the geometric details of (a) \& (b).
See particularly the last row of Figure~\ref{fig:shape_mixing};
the complex armrests and back in the input shape can be well preserved in our new shape, while
the chair's seat and legs are seamlessly connected with clean surfaces and sharp edges.

Further, we compare our method with SPAGHETTI~\cite{hertz2022spaghetti}, the state-of-the-art implicit method for the same task.
In particular, SPAGHETTI produces inverted shapes as a mixture of Gaussian distributions, where each of them is conditioned by a latent code.
Part replacement on the input can then be conducted by replacing some particular latent codes with the corresponding ones in another shape.
However, their results are 
somehow noisy 
in the connecting regions, while struggling to preserve the geometry and details of the input and part donor; see Figure~\ref{fig:shape_mixing} (c).
More visual comparisons on part replacement with SPAGHETTI~\cite{hertz2022spaghetti} and COALESCE~\cite{yin2020coalesce} are provided in supplementary material Section F.
Note also that for the examples shown in the main paper, we cannot provide comparisons of COALESCE, since it needs 
additional part-level
annotations, which are not available for these examples. 

\begin{figure}[t]
	\centering
	\includegraphics[width=0.9\linewidth]{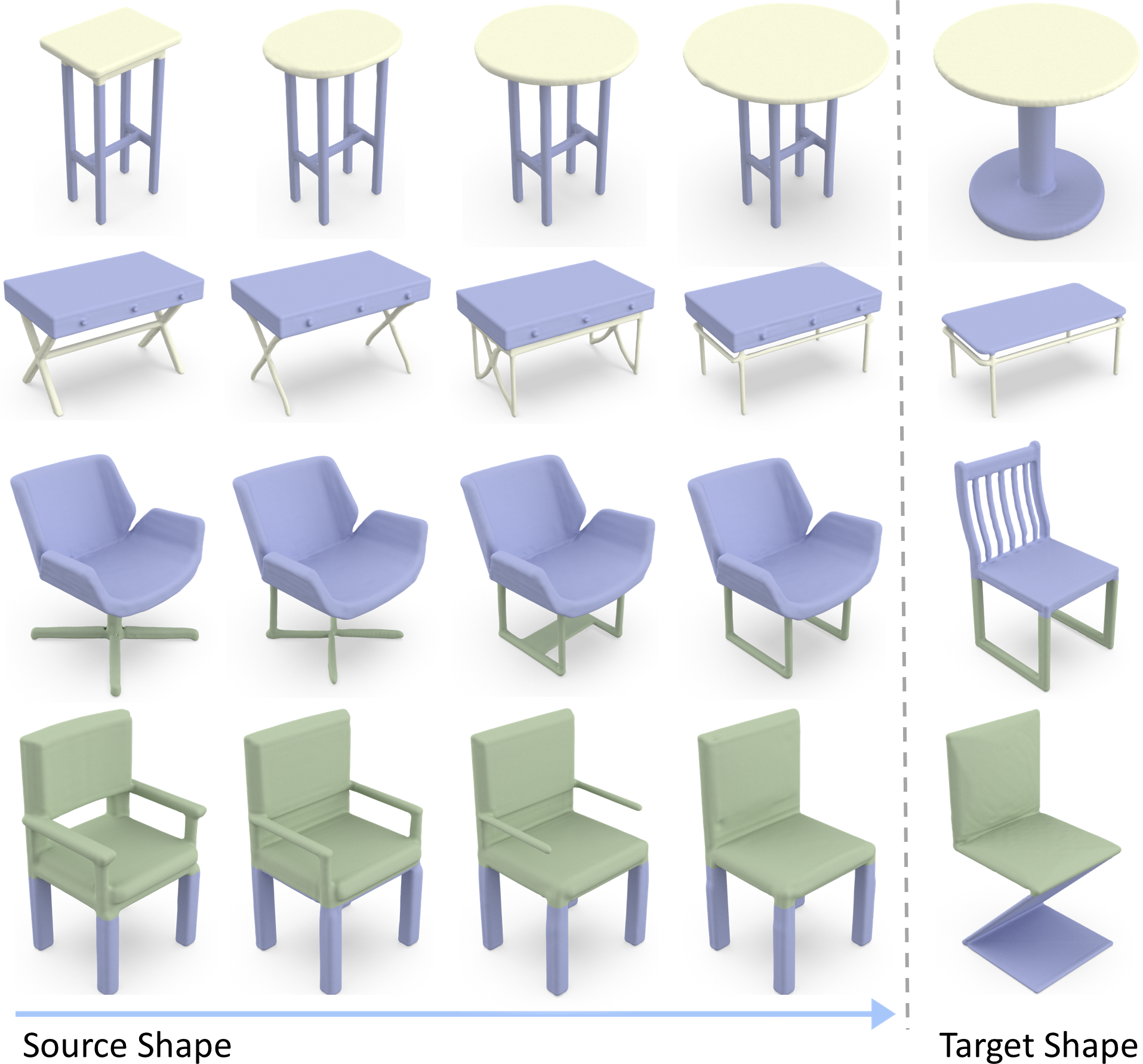}
	\vspace{-2mm}
	\caption{Part-wise interpolation results.
 Our method can interpolate the selected part (yellow for the tables and green for the chairs) in the sources (leftmost) towards the targets (rightmost). As shown, the intermediate results morph smoothly on the selected region, while preserving other parts.}
	\label{fig:part_interpolation}
	\vspace{-2mm}
\end{figure}

\vspace{-3pt}
\paragraph{Part-wise Interpolation.}
Also, our method enables us to select a part (region) in the source shape and perform interpolation on it towards the target shape.
See,~\eg, Figure~\ref{fig:part_interpolation};
during the interpolation, the green part of the source  morphs smoothly towards the associated part in the target shape while preserving the remaining parts (marked in blue).
See the drawer's knobs of the table in the second row and the chair's legs in the third row of Figure~\ref{fig:part_interpolation}, the geometry and topology of the untouched parts are preserved faithfully.
Besides, the selected part of the intermediate results is consistent with the untouched part; see the connecting regions between the chair's seat and legs in the last row of Figure~\ref{fig:part_interpolation}.

\begin{figure}[t]
	\centering
	\includegraphics[width=0.95\linewidth]{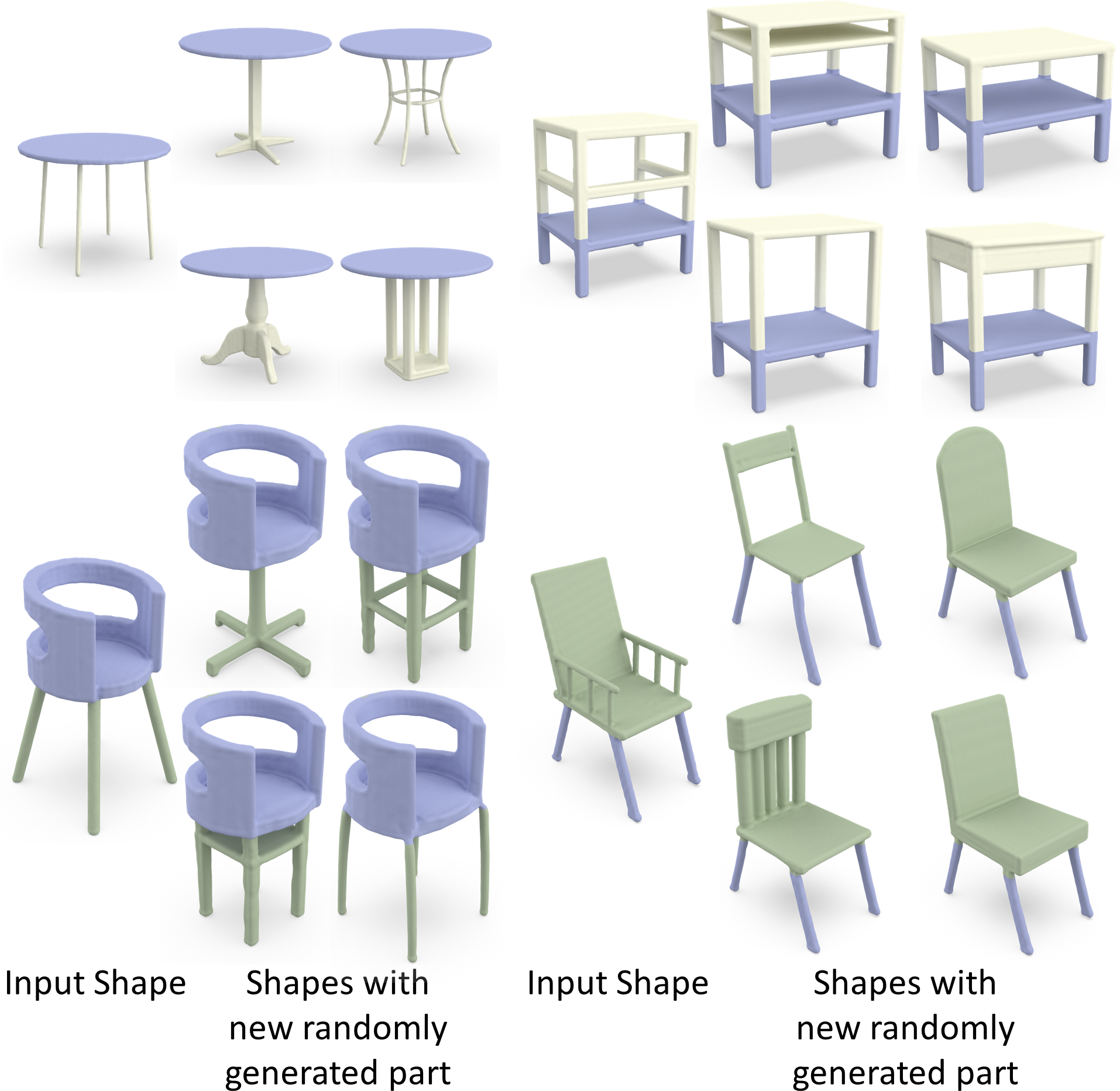}
	\vspace{-2mm}
	\caption{Part-wise re-generation results. We keep the blue parts fixed and randomly re-generate the remaining parts. Note the diverse structures and geometries in the re-generated parts. See the chair at bottom right; new parts replace the original back and its seat can exhibit various styles,~\eg, one with vertical bars and the other with a round back. Also, the geometries of the legs are consistent across the results.}
	\label{fig:part_generation}
	\vspace{-2mm}
\end{figure}

\vspace{-3pt}
\paragraph{Part-wise Re-generation.}
Further, our method allows us to specify a part in the input shape and replace it with a randomly-generated part; see Figure~\ref{fig:part_generation}.
The randomly-generated parts are diverse and plausible; see the table at top left of Figure~\ref{fig:part_generation}. The connecting regions in the tables keep consistent across all the re-generation results, while the re-generated legs of the tables exhibit diverse topology structures and geometric details.

\vspace{-3pt}
\paragraph{More shape manipulation results.}
Please refer to supplementary material Section F for more results on shape interpolation, part replacement, part-wise interpolation, and part-wise re-generation.


\vspace{-3pt}
\subsection{Shape Reconstruction}
\paragraph{Shape reconstruction from point clouds and single-view images}
By utilizing the latent space learned by the encoder network, we can leverage our method to help reconstruct implicit shapes from various forms of inputs.
As suggested by~\cite{chen2019learning}, we propose to train another encoder that takes a point cloud or a single-view image as input and to predict a latent code that matches the one obtained by our trained encoder network in Section~\ref{ssec:shape_learning}.
Given an unseen input (a point cloud or a single-view image), we can then use the new encoder to generate a corresponding latent code and take the code to reconstruct the 3D shape.

\begin{figure}[t]
	\centering
	\includegraphics[width=0.95\linewidth]{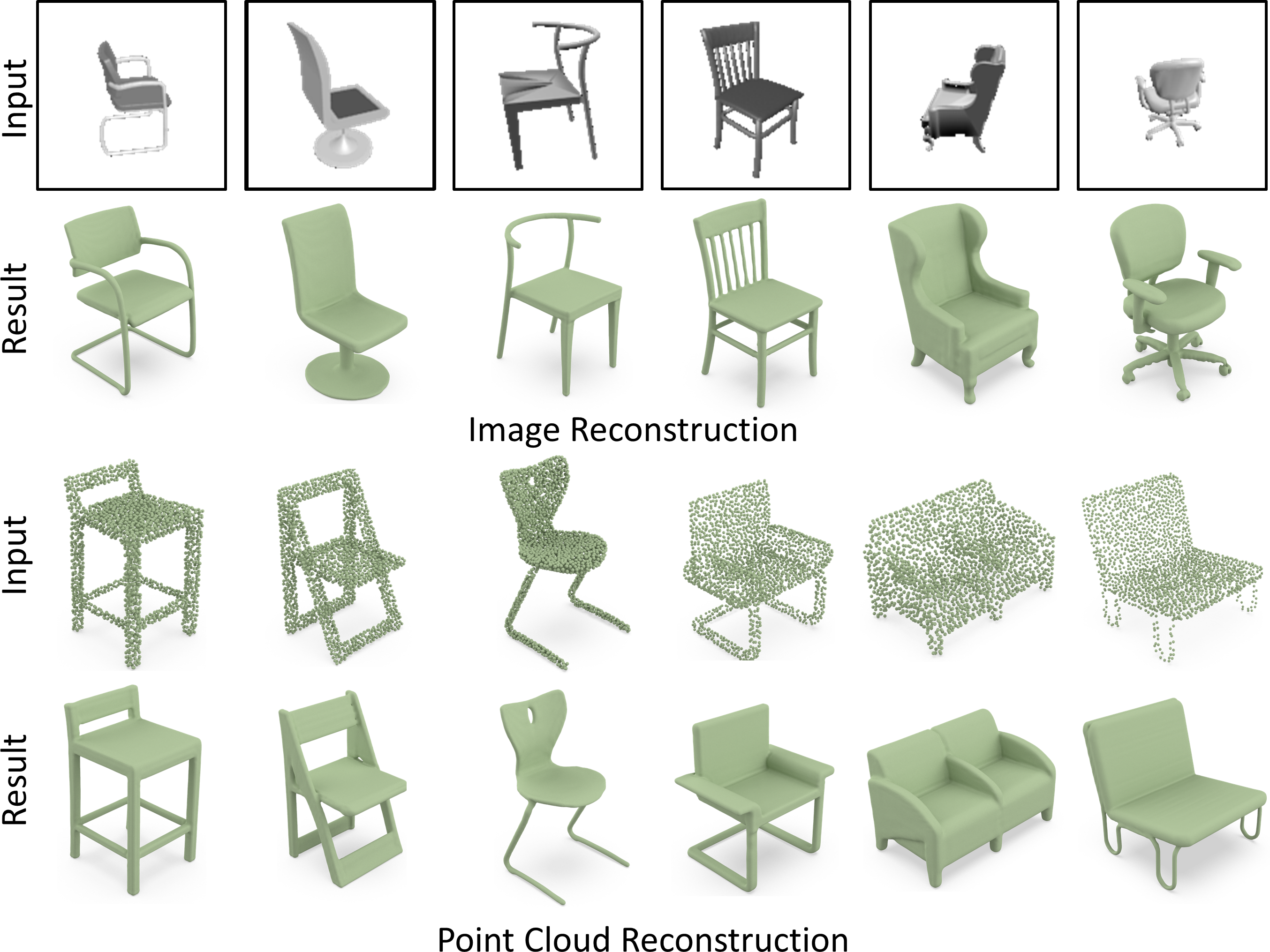}
	\vspace{-2mm}
	\caption{Shape reconstruction results.
 Our method can embed single-view images (top) and point clouds (bottom) into our latent space and faithfully reconstruct shapes that match the raw inputs.
 Note the complex topology structures and fine geometric details in the reconstructed shapes.}
	\label{fig:img_recon}
	\vspace{-2mm}
\end{figure}

Figure~\ref{fig:img_recon} shows visual examples of our reconstruction results.
We can reconstruct shapes with fine details and thin structures solely by predicting the latent codes.
The results show that the learned latent space is highly smooth and covers various unseen inputs.
Interestingly, our method can reconstruct some occluded parts in the single-view images,~\eg, the occluded legs of the chairs, by leveraging the shape priors learned in the latent space.

\begin{figure}
	\includegraphics[width=0.98\linewidth]{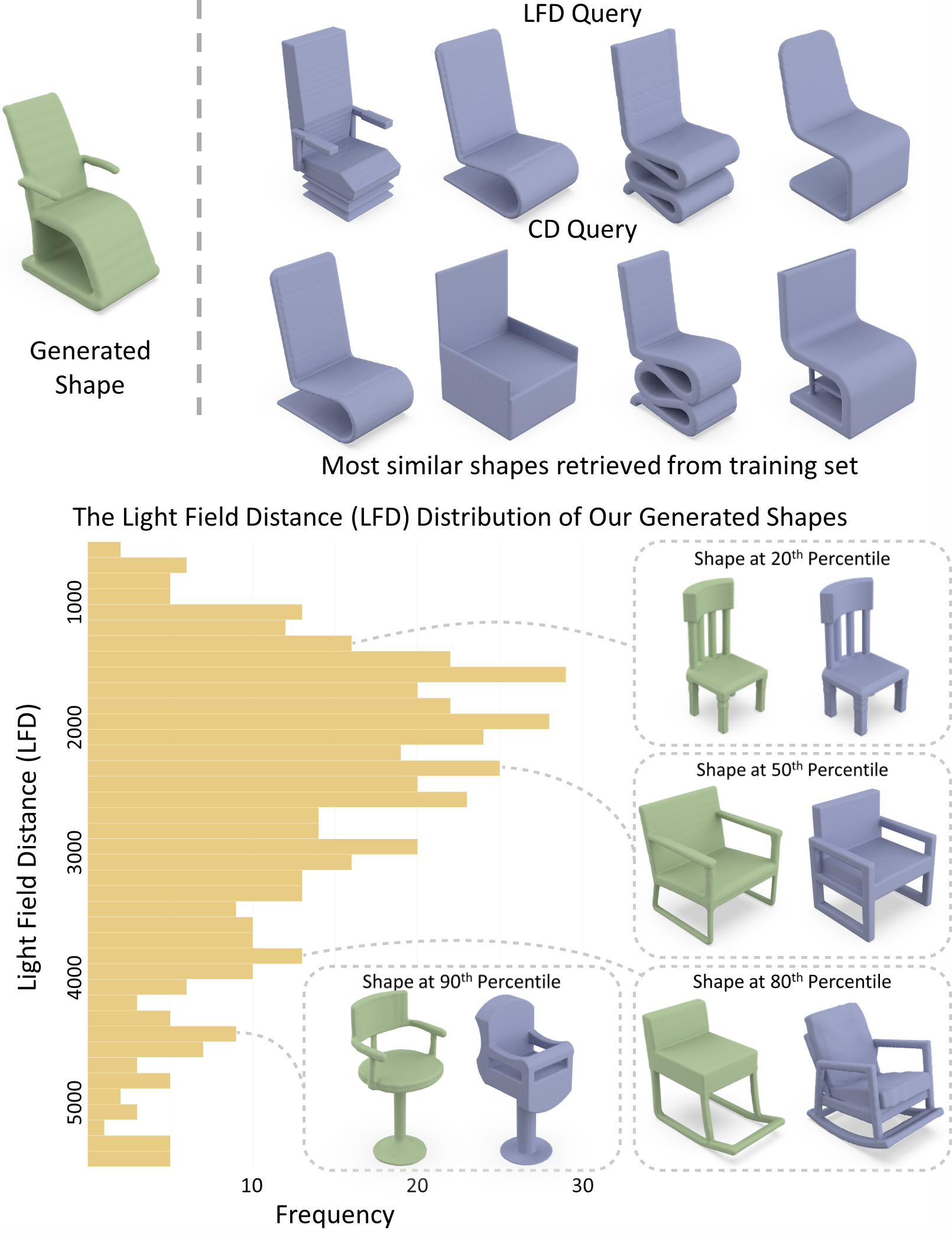}
	\vspace{-3mm}
\caption{
{
Shape novelty analysis.
Top: From our generated shape (in green),
we retrieve top-four most similar shapes (in blue) in training set by CD and LFD.
Bottom: We generate 500 chairs using our method;
for each chair, we retrieve the most similar shape in the training set by LFD; then, we plot the distribution of LFDs for all retrievals, showing that our method is able to generate shapes that are more similar (low LFDs) or more novel (high LFDs) compared to the training set.
Note that the generated shape at $50^{\text{th}}$ percentile is already not that similar to the associated training-set shape.
}
	\vspace{-1mm}
}
	\label{fig:novelty_analysis}
\end{figure}

\vspace*{-3pt}
\subsection{Novelty analysis on Shape Generation}
Next, we analyze whether our method can generate shapes that are not necessarily the same as the training-set shapes, meaning that it does not simply memorize the training data.
To do so, we use our method to generate 500 random shapes
and retrieve top-four most similar shapes in the training set separately via two different metrics,~\ie, Chamfer Distance (CD) and Light Field Distance (LFD)~\cite{chen2003visual}.
It is noted that LFD is computed based on rendered images from multiple views on each shape, so it focuses more on the visual similarity between shapes and is considered to be more robust for shape retrieval.
For the details on the metrics, please see supplementary material Section H.

Figure~\ref{fig:novelty_analysis} (top) shows a shape generated by our method, together with top-four most similar shapes retrieved from the training set by the CD and LFD metrics.
Further, we show another ten
examples in supplementary material Section G.
Comparing our shapes with the retrieved ones,
we can see that the shapes share similar structures, showing that our method is able to generate realistic-looking structures like those in the training set.
Beyond that, our shapes exhibit noticeable differences in various local structures.

As mentioned earlier, a good generator should produce diverse shapes that are not necessarily the same as the training shapes. 
So, we further statistically analyze the novelty of our generated shapes relative to the training set.
To do so, we use our method to generate 500 random chairs; for each generated chair shape, we use LFD to retrieve the most similar shape in the training set.
Figure~\ref{fig:novelty_analysis} (bottom) plots the distribution of LFDs between our generated shapes (in green) and retrieved shapes (in blue).
Also, we show four shape pairs at various percentiles, revealing that shapes with larger LFDs are more different from the most similar shapes in the training set.
From the LFD distribution, we can see that our method can learn a generation distribution that covers shapes in the training set (low LFD) and also generates novel and realistic-looking shapes that are more different (high LFD) from the training-set shapes.

\vspace*{-3pt}
\subsection{Ablation Study}

\paragraph{Ablation on shape generation.}
To evaluate the major components in shape generation, we 
successively ablate its full pipeline.
First, we evaluate the generation performance with/without the detail predictor.
Next, we study the contributions of the diffusion model and the wavelet representation in the generator network.

\input{tables/analysis}

From the results reported in Table 2, we can see the capability of the detail predictor, which introduces a substantial improvement on all metrics (first vs. second rows).
Further, replacing our generator with the VAD model or directly predicting TSDF leads to a performance degradation (second \& last two rows). %

\input{tables/inv_ablation_table.tex}

\vspace{-3pt}
\paragraph{Ablation on shape inversion.}
Next, we explore our shape inversion pipeline on the Chair category to evaluate its major components.
First, we quantitatively evaluate the effect of directly reconstructing the input from the latent code predicted by the encoder without the shape-guided refinement.
Second, we evaluate the shape inversion performance with/without the detail predictor.
Table~\ref{tab:inv_ablation} reports the quantitative results, demonstrating the capability of shape-guided refinement (first vs. second rows).
Further, dropping the detail predictor degrades the performance (first vs. third rows).
\begin{figure}[t]
	\centering
	\includegraphics[width=1.0\linewidth]{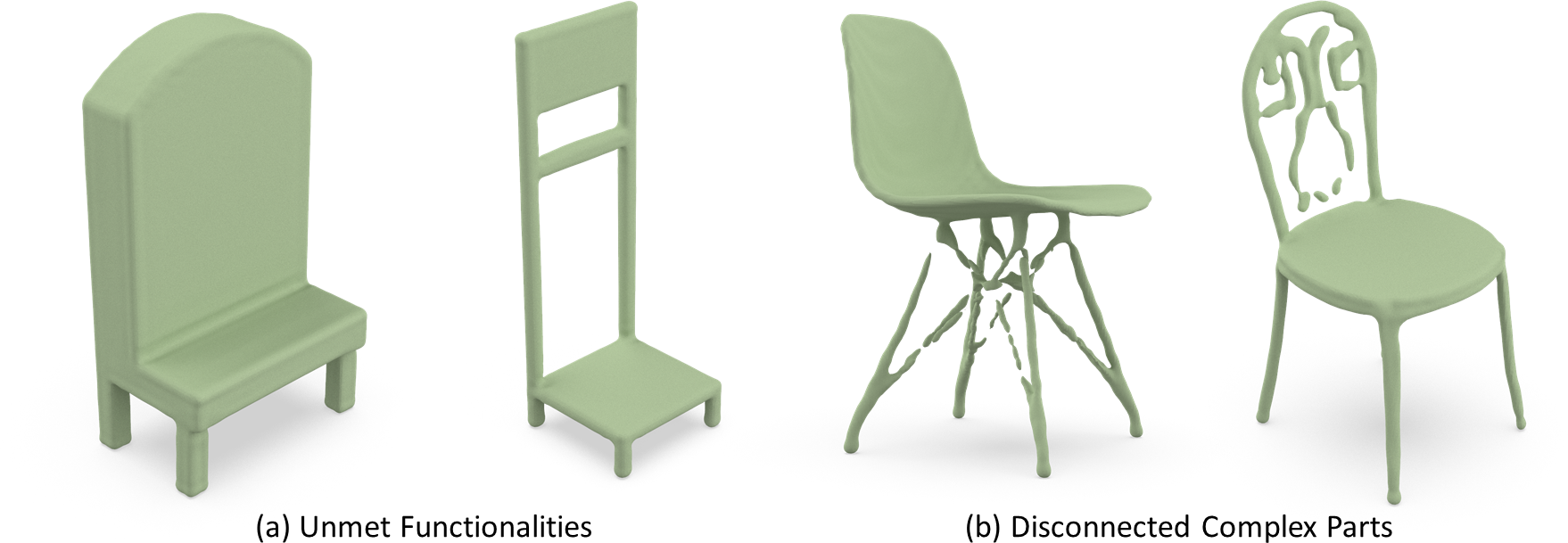}
	\caption{Failure cases.
	Left: chairs that are unlikely to meet the basic functionality in real world.
	Right: artifacts in complex and very thin structures.}
	\label{fig:fail_cases}
\end{figure} 

\vspace*{-3pt}
\paragraph{Details of ablation study}
Please refer to supplementary material Section M for details on how the ablation cases are implemented and visual comparison results on shape generation and inversion.

\vspace{-3pt}
\subsection{Limitations and Future works.}

\paragraph{Discussion on shape generation.}
While our method can generate diverse and realistic-looking shapes, the generated shapes may not meet the desired functionality.
As Figure~\ref{fig:fail_cases}(a) shows, the generated chairs, despite
looking interesting and structurally reasonable, have an exceptionally tall seat back and low seat height for normal human bodies.
In the future, we may incorporate functionality,~\eg,~\cite{blinn2021learning}, into the shape generation process.
Second, though our method can learn efficiently via the wavelet representation, it still requires a large number of shapes for training. 
So, for 
categories with few training samples, especially those with complex and very thin structures; see,~\eg, Figure~\ref{fig:fail_cases}(b), the generated shape may exhibit artifacts like broken parts and structures.
Last, while our method can better fit the data distribution than existing methods, the generated shapes still conform to structures/appearances in the training set. Exploring how to generate out-of-distribution shapes
is an interesting but challenging future direction.

\vspace{-3pt}
\paragraph{Discussion on shape inversion and manipulation.}
Our method is able to invert and reconstruct shapes faithfully and facilitate manipulations with high fidelity, yet requiring users to manually select regions for manipulation.
An interesting future direction is to guide the shape manipulation by sketches or texts~\cite{li2017bendsketch, liu2022towards}.
Also, while our compact wavelet representation maintains a good spatial correspondence with the original TSDF domain and enables various part-aware manipulations, inter-shape correspondences still 
rely
mainly on a canonically-aligned dataset.
We hope to establish stronger correspondences in the compact wavelet domain to enable more compelling downstream applications such as texture and detail transfer.
Besides, the diffusion process requires a large number of 
time steps, so producing manipulated shapes is 
time-consuming.
Yet, simply reducing the time steps leads to a significant quality drop.
We plan to explore acceleration strategies to promote interactive shape manipulation.

%% file: tables/inversion_quantitative.tex
\begin{table*}[t]
	\centering
		\caption{
		Quantitative comparison on shape inversion between our method and four state-of-the-art methods.
		We follow the same setting as in the state-of-the-art methods.
        Our method outperforms (lowest CD, EMD, and LFD) all the other methods for all categories.
        The units of CD, EMD, and LFD are $10^{-3}$, $10^{-2}$, and $1$, respectively.
        Notice that the quantitative evaluation for SPAGHETTI~\cite{hertz2022spaghetti} in the Lamp category is not shown here, as SPAGHETTI does not provide the pre-trained model of Lamp and the training code has not been officially released.}
        \vspace*{-2mm}
	\resizebox{0.82\linewidth}{!}{
		\begin{tabular}{C{5cm}|C{0.6cm}C{0.6cm}C{0.9cm}|C{0.6cm}C{0.6cm}C{0.9cm}|C{0.6cm}C{0.6cm}C{0.9cm}}
			\toprule[1pt]
                        \multirow{2}*{Method} & \multicolumn{3}{c|}{Chair}                                               & \multicolumn{3}{c|}{Airplane}& \multicolumn{3}{c}{Lamp}
                        \\ 
                         & CD$~\downarrow$        & EMD$~\downarrow$        & LFD$~\downarrow$         & CD$~\downarrow$        & EMD$~\downarrow$         & LFD$~\downarrow$  & CD$~\downarrow$        & EMD$~\downarrow$         & LFD$~\downarrow$ \\ \hline
IM-NET~\cite{chen2019learning}
& 4.968 &  10.42 & 2800.56 & 3.301 &  9.261 & 5024.65  & 20.253 & 20.403 & 6880.01
\\ \hline
DeepSDF~\cite{kleineberg2020adversarial}
&  4.657 & 9.066 & 2403.08 & 3.249 & 9.734 & 4939.56  & 17.540 & 18.918 & 6858.34
\\ \hline
DualSDF~\cite{luo2021diffusion}
&  8.254 & 13.01 & 2588.87 & 6.529 & 15.74 & 7097.09 & 35.125 & 26.692 & 6820.62
\\ \hline
SPAGHETTI~\cite{hertz2022spaghetti}
& 2.837 & 7.663 & 1988.82 & 1.386 & 6.466 & 3637.33  &  ------ & ------ &  ------
\\ \hline \hline

Ours without Refinement & 4.578 & 8.919 & 2358.98 & 2.459 & 6.816 & 3726.43 & 17.412 & 19.237 & 6955.02  \\
Ours  &  \textbf{1.142} & \textbf{4.491} & \textbf{924.911} & \textbf{1.108} & \textbf{4.507} & \textbf{2508.55} & \textbf{5.834} & \textbf{7.201} & \textbf{3450.34}  \\

			\bottomrule[1pt]
	\end{tabular}}
    \vspace*{-1mm}
\label{tab:inv_quanComparison}
\end{table*}

%% file: tables/analysis.tex
\begin{table}[t]
	\centering
		\caption{Comparing our full generation pipeline with various ablated cases on the Chair category.
		The unit of CD is $10^{-3}$ and the unit of EMD is $10^{-2}$.
  }	
		\vspace*{-2mm}
		\resizebox{0.95\linewidth}{!}{
		\begin{tabular}{C{3cm}|C{0.5cm}C{0.5cm}C{0.5cm}C{0.5cm}C{0.5cm}C{0.5cm}}
			\toprule[1pt]
			\multirow{2}{*}{Method} 
			& \multicolumn{2}{c}{COV $\uparrow$}
			& \multicolumn{2}{c}{MMD $\downarrow$}
			& \multicolumn{2}{c}{1-NNA $\sim\hspace*{-1mm}50$} \\
			& CD & EMD & CD & EMD & CD & EMD \\ \hline
			Full Model & \textbf{58.19} & \textbf{55.46} & \textbf{11.70} & \textbf{14.31} & \textbf{61.47} & \textbf{61.62} \\ \hline
			W/o detail predictor & 54.20 & 50.96 & 12.32 & 14.54 & 62.46 & 62.57 \\
			VAD Generator & 21.83 & 26.77 & 21.83 & 26.77 & 95.20 & 93.62 \\
			 Direct predict TSDF & 50.51 & 50.67 & 12.83 &15.24 & 68.69 & 68.29 \\
			\bottomrule[1pt]
	\end{tabular}}
	\label{tab:analysis}
	\vspace{-2mm}
\end{table} 

%% file: tables/inv_ablation_table.tex
\begin{table}[t]
	\centering
 \caption{Comparing our full inversion pipeline with various ablated cases on the Chair category. The units of CD, EMD, and LFD are $10^{-3}$, $10^{-2}$, and 1, respectively.}

		\vspace*{-2mm}
		\resizebox{0.95\linewidth}{!}{
		\begin{tabular}{C{4.0cm}|C{1.0cm}C{1.0cm}C{1.0cm}}
			\toprule[1pt]
			\multirow{1}{*}{Method} 
			& \multicolumn{1}{c}{CD $\downarrow$ }
			& \multicolumn{1}{c}{EMD $\downarrow$ }
			& \multicolumn{1}{c}{LFD $\downarrow$ } \\
                \hline
			Full Model & \textbf{1.142} & \textbf{4.491} & \textbf{924.91}  \\ \hline
			W/o Refinement & 4.578 & 8.919 & 2358.98  \\

			W/o detail predictor & 1.194 & 5.129 & 1200.04  \\
			\bottomrule[1pt]
	\end{tabular}}
	\label{tab:inv_ablation}
	\vspace{-2mm}
\end{table}

%% file: conclusion.tex
\section{Conclusion}

This paper presents a new framework for 3D shape generation, inversion, and manipulation.
Unlike prior works, we operate on the frequency domain.
By decomposing the implicit function in the form of the TSDF using biorthogonal wavelets, 
we build a compact wavelet representation with a pair of coarse and detail coefficient volumes, as an encoding of 3D shape.
Then, we formulate our generator upon
a probabilistic diffusion model to learn to generate diverse shapes in the form of coarse coefficient volumes from noise samples and a detail predictor to further learn to generate compatible detail coefficient volumes for reconstructing fine details.
Further, by introducing an encoder into the generation process, our framework can enable faithful inversion of unseen shapes, shape interpolation, and a rich variety of region-aware manipulations.
Both quantitative and qualitative experiments show superior capabilities of our new approach on shape generation, inversion, and manipulation over the state-of-the-art methods.